\documentclass[sigconf]{acmart} 

\usepackage{booktabs} 

\usepackage{microtype}
\usepackage{graphicx}
\usepackage{subfigure}
\usepackage{booktabs} 
\usepackage{makecell}
\usepackage{hyperref}
\usepackage{pdflscape}
\usepackage{mathrsfs}
\usepackage{paralist}
\usepackage{multirow}
\usepackage{enumitem}
\usepackage{amsmath}
\usepackage{bbold}
\usepackage{mathtools}
\usepackage[normalem]{ulem}
\usepackage{amsthm}

\newcommand{\bit}{\begin{compactitem}}
\newcommand{\eit}{\end{compactitem}}
\newcommand{\ben}{\begin{compactenum}}
\newcommand{\een}{\end{compactenum}}
\newcommand{\mytag}[1]{{\bf#1}}

\newcommand{\modelTitle}{{\bf MoFlow\ }}
\newcommand{\model}{MoFlow\ }
\newcommand{\modelEOL}{MoFlow}

\newcommand{\hide}[1]{}

\DeclareMathOperator*{\argmax}{arg\,max}

\usepackage[ruled,vlined]{algorithm2e}

\theoremstyle{definition}
\newtheorem{definition}{Definition}[section]
\theoremstyle{remark}

\setcopyright{acmlicensed}

\copyrightyear{2020}
\acmYear{2020}
\setcopyright{acmlicensed}\acmConference[KDD '20]{Proceedings of the 26th ACM SIGKDD Conference on Knowledge Discovery and Data Mining}{August 23--27, 2020}{Virtual Event, CA, USA}
\acmBooktitle{Proceedings of the 26th ACM SIGKDD Conference on Knowledge Discovery and Data Mining (KDD '20), August 23--27, 2020, Virtual Event, CA, USA}
\acmPrice{15.00}
\acmDOI{10.1145/3394486.3403104}
\acmISBN{978-1-4503-7998-4/20/08}
\hide{
\copyrightyear{2019} 
\acmYear{2019} 
\setcopyright{acmcopyright}
\acmConference[KDD '19]{The 25th ACM SIGKDD Conference on Knowledge Discovery and Data Mining}{August 4--8, 2019}{Anchorage, AK, USA}
\acmBooktitle{The 25th ACM SIGKDD Conference on Knowledge Discovery and Data Mining (KDD '19), August 4--8, 2019, Anchorage, AK, USA}
\acmPrice{15.00}
\acmDOI{10.1145/3292500.3330842}
\acmISBN{978-1-4503-6201-6/19/08}
}

\begin{document}
\title{
MoFlow: An Invertible Flow Model for Generating\\ Molecular Graphs}
 
\author{Chengxi Zang}
\affiliation{%
  \institution{Department of Population Health Sciences, Weill Cornell Medicine
}
}
\email{chz4001@med.cornell.edu}

\author{Fei Wang}
\affiliation{%
  \institution{Department of Population Health Sciences, Weill Cornell Medicine 
}
}
\email{few2001@med.cornell.edu}



\begin{abstract}
Generating molecular graphs with desired chemical properties driven by deep graph generative models provides a very promising way to accelerate 
 drug discovery process. 
Such graph generative models usually consist of two steps: learning latent representations and generation of molecular graphs. However, to generate novel and chemically-valid molecular graphs from latent representations is very challenging because of  the chemical constraints and combinatorial complexity of molecular graphs. In this paper, we propose \modelEOL, a flow-based graph generative model to learn invertible mappings between molecular graphs and their latent representations.
To generate molecular graphs, our \model  first generates bonds (edges) through a Glow based model, then generates atoms (nodes) given bonds by a novel graph conditional flow, and finally assembles them into a chemically valid molecular graph with a posthoc validity correction. Our \model has merits including exact and tractable likelihood training, efficient one-pass embedding and generation, chemical validity guarantees, 100\% reconstruction of training data, and good generalization ability. We validate our model by four tasks: molecular graph generation and reconstruction, visualization of the continuous latent space, property optimization, and constrained property optimization. Our \model achieves state-of-the-art performance, which implies its potential efficiency and effectiveness to explore large chemical space for drug discovery.

\end{abstract}

%
%




\maketitle

\section{Introduction}
\label{sec:intro}

Drug discovery aims at finding candidate molecules with desired chemical properties for clinical trials, which is a long (10-20 years) and costly (\$0.5-\$2.6 billion) process with a high failure rate \cite{paul2010improve,avorn20152}. Recently, deep graph generative models have demonstrated their big potential to accelerate the drug discovery process by exploring large chemical space in a data-driven manner \cite{jin2018junction,zhavoronkov2019deep}. These models usually first learn a continuous latent space by encoding\footnote{In this paper, we use inference, embedding or encoding interchangeably to refer to the transformation from molecular graphs to the learned latent space, and we use decoding or generation for the reverse transformation.} the training molecular graphs and then generate novel and optimized ones through decoding from the learned latent space guided by targeted properties \cite{gomez2018automatic,jin2018junction}. 
However, it is still very challenging to generate novel and chemically-valid molecular graphs with desired properties since: a) the scale of the chemical space of drug-like compounds is $10^{60}$ \cite{mullard2017drug} but the scale of possibly generated molecular graphs by existing methods are much smaller, and
b) generating molecular graphs that have both multi-type nodes and edges  and follow bond-valence constraints
is a hard combinatorial task.

Prior works leverage different deep generative frameworks for generating molecular SMILES codes \cite{weininger1989smiles} or molecular graphs, including variational autoencoder (VAE)-based models \cite{kusner2017grammar,dai2018syntax,simonovsky2018graphvae,ma2018constrained,liu2018constrained,bresson2019two,jin2018junction}, generative adversarial networks (GAN)-based models \cite{de2018molgan,you2018graph}, and autoregressive (AR)-based models \cite{popova2019molecularrnn,you2018graph}. 
In this paper, we explore a different deep generative framework, namely the normalizing flow \cite{dinh2014nice,madhawa2019graphnvp,kingma2018glow} to generate molecular graphs. Compared with above three frameworks, the flow-based models are the only one which can memorize and exactly reconstruct all the input data, and at the same time have the potential to generate more novel, unique and valid molecules, which implies its potential capability of deeper exploration of the huge chemical space. To our best knowledge, there have been three flow-based models proposed for molecular graph generation. The GraphAF \cite{shi2020graphaf} model is an autoregressive flow-based model that achieves state-of-the-art performance in molecular graph generation. GraphAF generates molecules in a sequential manner by adding each new atom or bond followed by a validity check. GraphNVP \cite{madhawa2019graphnvp} and GRF \cite{honda2019graph} are proposed for molecular graph generation in a one-shot manner. However, they cannot guarantee chemical validity and thus show poor performance in generating  valid and novel molecules. 

In this paper, we propose a novel deep graph generative model named \model to generate molecular graphs.  Our \model is the first of its kind  which not only generates molecular graphs efficiently by invertible mapping at one shot, but also has a chemical validity guarantee. More specifically, to capture the combinatorial atom-and-bond structures of molecular graphs, we propose a variant of the Glow model \cite{kingma2018glow} to generate bonds (multi-type edges, e.g., single, double and triple bonds), a novel \textit{graph conditional  flow} to generate atoms (multi-type nodes, e.g. C, N etc.) given bonds 
by leveraging graph convolutions, and finally assemble atoms and bonds  into a valid molecular graph which follows bond-valence constraints.
We illustrate our modelling framework in Figure~\ref{fig:model}. Our \model is trained by exact and tractable likelihood estimation, and one-pass inference and generation can be efficiently utilized for molecular graph optimization.

We validate our \model through a wide range of experiments from molecular graph generation, reconstruction, visualization to optimization. As baselines, we compare the state-of-the-art VAE-based model \cite{jin2018junction}, autoregressive-based models \cite{you2018graph,popova2019molecularrnn}, and all three flow-based models \cite{madhawa2019graphnvp,honda2019graph, shi2020graphaf}. 
As for memorizing input data, \model achieves $100\%$ reconstruction rate. As for exploring the unknown chemical space, \model outperforms above models by generating more novel, unique and valid molecules (as demonstrated by the N.U.V. scores in Table~\ref{tab:qm9} and~\ref{tab:zinc}). \model generates $100\%$ chemically-valid molecules when sampling from prior distributions. Furthermore, if without validity correction, \model still generates much more valid molecules than existing models (validity-without-check scores in Table~\ref{tab:qm9} and~\ref{tab:zinc}). For example, the state-of-the-art autoregressive-flow-based model GraphAF \cite{shi2020graphaf} achieves $67\%$ and $68\%$ validity-without-check scores for two datasets while \model achieves $96\%$ and $82\%$ respectively, thanks to its capability of capturing the chemical structures in a holistic way.
As for chemical property optimization, \model can find much more novel molecules with top drug-likeness scores than existing models (Table~\ref{tab:topqed} and Figure~\ref{fig:topqed}). As for constrained property optimization, \model finds novel and optimized molecules with the best similarity scores and second best property improvement (Table~\ref{tab:plogp}). 

It is worthwhile to highlight our contributions as follows:
  \bit
 \item {\bf Novel \model model:}
  our \model is one of the first flow-based graph generative models which not only generates molecular graphs at one shot by invertible mapping but also has a validity guarantee. To capture the combinatorial atom-and-bond structures of molecular graphs, we propose a variant of Glow model for bonds (edges) and a novel  graph conditional  flow for atoms (nodes) given bonds, and then assemble them into valid molecular graphs. 
 
\item {\bf State-of-the-art performance:} our \model achieves many state-of-the-art results w.r.t. molecular graph generation, reconstruction, optimization, etc., and at the same time our one-shot inference and generation are very efficient, which implies its potentials in deep exploration of huge chemical space for drug discovery.
\eit

The outline of this paper is: 
survey (Sec.~\ref{sec:related}), proposed method (Sec.~\ref{sec:mp} and \ref{sec:model}), experiments (Sec.~\ref{sec:exp}), and conclusions (Sec.~\ref{sec:conclusion}). In order to promote reproducibility, our codes and datasets are open-sourced  at \url{https://github.com/calvin-zcx/moflow}.

\section{Related Work}
\label{sec:related}
\mytag{Molecular Generation.}
Different deep generative frameworks are proposed for generating molecular SMILES or molecular graphs. Among the variational autoencoder (VAE)-based models \cite{kusner2017grammar,dai2018syntax,simonovsky2018graphvae,ma2018constrained,liu2018constrained,bresson2019two,jin2018junction}, the JT-VAE \cite{jin2018junction} generates valid tree-structured molecules by first generating a tree-structured scaffold of chemical substructures and then assembling substructures according to the generated scaffold.
The MolGAN \cite{de2018molgan} is a generative adversarial networks (GAN)-based model but shows very limited performance in generating valid and unique molecules. The autoregressive-based models generate molecules in a sequential manner with validity check at each generation step. For example, the MolecularRNN \cite{popova2019molecularrnn} sequentially generates each character of SMILES and the GCPN \cite{you2018graph} sequentially generates each atom/bond in a molecular graphs.
In this paper, we explore a different deep generative framework, namely the normalizing flow models \cite{dinh2014nice,madhawa2019graphnvp,kingma2018glow}, for molecular graph generation, which have the potential to memorize and reconstruct all the training data and generalize to generating more valid, novel and unique molecules.


\noindent \mytag{Flow-based Models.}
The (normalizing) flow-based models try to learn mappings between complex distributions and simple prior distributions through invertible neural networks and such a framework has good merits of exact and tractable likelihood estimation for training, efficient one-pass inference and sampling, invertible mapping and thus reconstructing all the training data etc. Examples include NICE\cite{dinh2014nice}, RealNVP\cite{dinh2016density}, Glow\cite{kingma2018glow} and GNF \cite{liu2019graph} which show promising results in generating images or even graphs \cite{liu2019graph}. See latest reviews in \cite{papamakarios2019normalizing,kobyzev2019normalizing} and more technical details in Section~\ref{sec:mp}.


To our best knowledge, there are three flow-based models for molecular graph generation. The GraphAF \cite{shi2020graphaf} is an autoregressive flow-based model which achieves state-of-the-art performance in molecular graph generation. The GraphAF generates molecular graphs in a sequential manner with validity check when adding any new atom or bond. The GraphNVP \cite{madhawa2019graphnvp} and GRF \cite{honda2019graph} are proposed for molecular graph generation in a one-shot manner. However, they have no guarantee for chemical validity and thus show very limited performance in generating valid and novel molecular graphs. 
Our \model is the first of its kind which not only generates molecular graphs efficiently by invertible mapping at one shot but also has a validity guarantee. In order to capture the atom-and-bond composition of molecules, we propose a variant of Glow\cite{kingma2018glow}  model for bonds and a novel graph  conditional flow for atoms given bonds, and then combining them with a post-hoc validity correction. Our \model achieves many state-of-the-art results thanks to capturing the chemical structures in a holistic way, and  our one-shot inference and generation are more efficient than sequential models.
 



\section{Model Preliminary}
\label{sec:mp}
\mytag{The flow framework.} The flow-based models aim to learn a sequence of invertible transformations $f_{\Theta}=f_L \circ ... \circ f_1$ between complex high-dimensional data $X \sim P_\mathcal{X}(X)$ and  $Z \sim P_\mathcal{Z}(Z)$ in a latent space with the same number of dimensions where the latent distribution $P_\mathcal{Z}(Z)$ is easy to model (e.g., strong independence assumptions hold in such a latent space).  The potentially complex  data in the original space can be modelled by the \textbf{change of variable formula}  where $Z=f_{\Theta}(X)$ and:
\begin{equation}
\small
P_{\mathcal{X}}(X)=P_{\mathcal{Z}}(Z) \mid \det (\frac{\partial Z}{\partial X}) \mid.
\end{equation}
To sample $\tilde{X}\sim P_\mathcal{X}(X)$ is achieved by sampling $\tilde{Z} \sim P_\mathcal{Z}(Z)$ and then to transform $\tilde{X} = f_{\Theta}^{-1} (\tilde{Z})$ by the reverse mapping of $f_{\Theta}$.

Let $Z = f_{\Theta}(X) = f_L \circ ... \circ f_1 (X)$,  \ $H_l = f_l (H_{l-1})$ where $f_l$ ($l = 1, ... L \in \mathbb{N^+}$) are invertible mappings, $H_{0}=X$, $H_{L}=Z$ and  $P_\mathcal{Z}(Z)$ follows a standard isotropic Gaussian with independent dimensions. Then we get the log-likelihood of $X$ by the change of variable formula as follows:
\begin{equation}
\small
\begin{aligned}
\log P_{\mathcal{X}}(X) &=\log P_{\mathcal{Z}}(Z) + \log \mid \det (\frac{\partial Z}{\partial X}) \mid \\
&=\sum_i \log P_{\mathcal{Z}_i}(Z_i) + \sum_{l=1}^{L}\log \mid \det (\frac{\partial f_l}{\partial H_{l-1}}) \mid 
\end{aligned}
\end{equation} where $P_{\mathcal{Z}_i}(Z_i)$ is the probability of the $i^{th}$ dimension of $Z$ and $f_{\Theta}=f_L \circ ... \circ f_1$  is an invertible deep neural network  to be learnt. Thus, the exact-likelihood-based training is tractable.

\mytag{Invertible affine coupling layers.} However, how to design a.) an invertible function $f_{\Theta}$ with b.) expressive structures and c.) efficient computation of the Jacobian determinant are nontrivial. The NICE\cite{dinh2014nice} and RealNVP \cite{dinh2016density} design an affine coupling transformation $Z = f_{\Theta}(X):\mathbb{R}^{n} \mapsto \mathbb{R}^{n}$:
\begin{equation}
\small
\begin{aligned}
Z_{1:d} &= X_{1:d} \\
Z_{d+1:n} &= X_{d+1:n} \odot e^{S_{\Theta}(X_{1:d})} + T_{\Theta}(X_{1:d}),
\end{aligned}
\end{equation} by splitting $X$ into two partitions $X = (X_{1:d}, X_{d+1:n})$. Thus,  a.) the invertibility is guaranteed by:
\begin{equation}
\small
\begin{aligned}
X_{1:d} &= Z_{1:d} \\
X_{d+1:n} &= (Z_{d+1:n} - T_{\Theta}(Z_{1:d})) / e^{S_{\Theta}(Z_{1:d})},
\end{aligned}
\end{equation}
b.) the expressive power depends on arbitrary neural structures of the \textbf{{S}cale function}  $S_{\Theta}:\mathbb{R}^{d} \mapsto \mathbb{R}^{n-d}$ and the \textbf{{T}ransformation function} $T_{\Theta}:\mathbb{R}^{d} \mapsto \mathbb{R}^{n-d}$ in the affine transformation of $X_{d+1:n}$, and  c.) the Jacobian determiant can be computed efficiently by  $\det (\frac{\partial Z}{\partial X})=\exp{(\sum_j S_{\Theta}(X_{1:d})_j)}$. 

\mytag{Splitting Dimensions.} 
The flow-based models, e.g., RealNVP \cite{dinh2016density} and Glow \cite{kingma2018glow}, adopt
\mytag{squeeze operation} which compresses the spatial dimension $X^{c \times n \times n}$ into $X^{(ch^2) \times \frac{n}{h} \times \frac{n}{h}}$ to make more channels and then split channels into two halves for the coupling layer. A deep flow model at a specific layer transforms unchanged dimensions in the previous layer to  keep all the dimensions transformed. 
In order to learn an optimal partition of $X$, Glow \cite{kingma2018glow} model introduces an \mytag{ invertible $1 \times 1$ convolution} $: \mathbb{R}^{c \times n \times n} \times \mathbb{R}^{c \times c} \mapsto \mathbb{R}^{c \times n \times n}$
with learnable convolution kernel $W \in \mathbb{R}^{c \times c}$ which is initialized as a random rotation matrix. After the transformation $Y = \text{invertible $1 \times 1$ convolution}(X, W)$, a fixed partition $Y = (Y_{1:\frac{c}{2}, :, :}, Y_{\frac{c}{2}+1:n, :, :})$ over the channel $c$ is used for the affine coupling layers.




\mytag{Numerical stability by actnorm.}
In order to ensure the numerical stability of the flow-based models, \mytag{actnorm layer} is introduced in Glow \cite{kingma2018glow} which normalizes dimensions in each channel over a batch by an affine transformation with learnable scale and bias. The scale and the bias are initialized as the mean and the inverse of the standard variation of the dimensions in each channel over the batch.






\section{Proposed \modelTitle Model}
\label{sec:model}
\begin{figure}[!tb]
\vspace{-0.1in}
\centering
\includegraphics[width=0.52\textwidth]{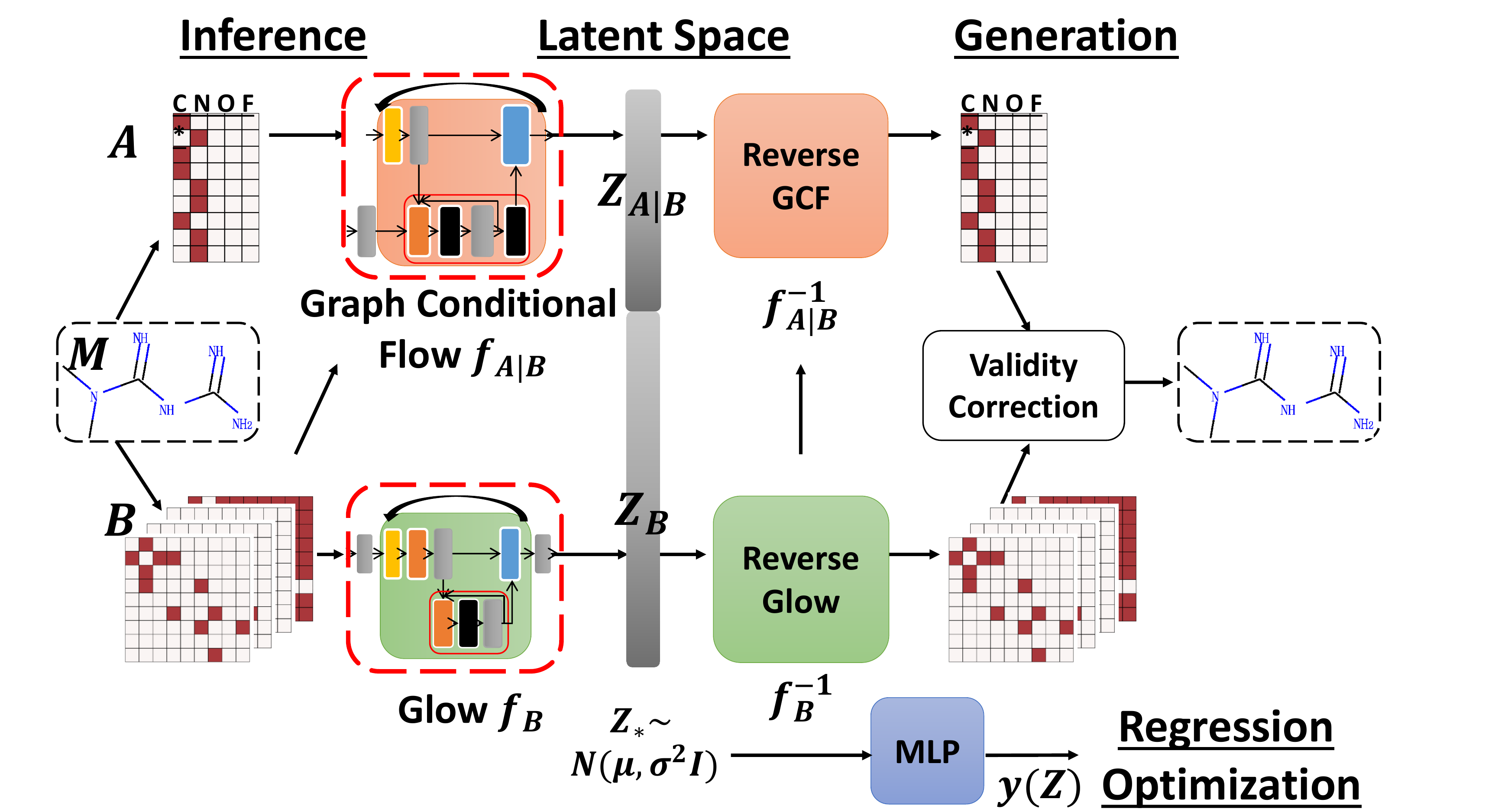}
\vspace{-0.2in}
\caption{ 
The outline of our \modelEOL. A molecular graph $M$ (e.g. Metformin) is represented by a feature matrix $A$ for atoms and adjacency tensors $B$ for bonds. Inference:
the graph conditional flow (GCF) $f_{\mathcal{A|B}}$ for atoms (Sec.~\ref{sec:model:gcf}) transforms $A$ given $B$ into conditional latent vector $Z_{A|B}$, and the Glow $f_{\mathcal{B}}$ for bonds (Sec.~\ref{sec:model:glow}) transform $B$ into latent vector $Z_B$. The latent space follows a spherical Gaussian distribution. Generation: the generation process is the reverse transformations of previous operations, followed by a validity correction (Sec.~\ref{sec:model:validity}) procedure which ensures the chemical validity. We summarize \model in  Sec.~\ref{sec:model:all}. Regression and optimization: 
 the mapping  $y(Z)$ between latent space and molecular properties are used for
molecular graph optimization and property prediction
(Sec.~\ref{sec:exp:opt}, Sec.~\ref{sec:exp:copt}).
\label{fig:model}}
\vspace{-0.1in}
\end{figure}

In this section, we first define the problem and then introduce our \underline{Mo}lecular \underline{Flow}  (\modelEOL) model in detail. We show the outline of our \model in Figure~\ref{fig:model} as a roadmap for this section.

\subsection{Problem Definition: Learning a Probability Model of Molecular Graphs}
\label{sec:model:problem}
Let $\mathcal{M}=\mathcal{A} \times \mathcal{B} \subset  \mathbb{R}^{n \times k} \times \mathbb{R}^{c \times n \times n}$ denote the set of $\mathcal{M}$olecules which is the Cartesian product of the $\mathcal{A}$tom set $\mathcal{A}$ with at most $n \in \mathbb{N^+}$ atoms belonging to $k \in \mathbb{N^+}$ atom types and the $\mathcal{B}$ond set $\mathcal{B}$ with $c\in \mathbb{N^+}$ bond types.  A molecule $M=(A,B) \in \mathcal{A} \times \mathcal{B}$ is a pair of an atom matrix $A \in \mathbb{R}^{n \times k}$ and a bond tensor $B \in \mathbb{R}^{c \times n \times n}$.
We use one-hot encoding for the empirical molecule data where
$A (i, k) = 1$ represents the atom $i$ has atom type $k$, and $B (c, i, j)=B (c, j, i)=1$ represents a type $c$ bond between atom $i$ and atom $j$. Thus, a molecule $M$ can be viewed as an undirected graph with multi-type nodes and multi-type edges.

Our primary goal is to learn a molecule generative model $P_{\mathcal{M}} (M)$ which is the probability of sampling any molecule  $M$ from $P_{\mathcal{M}}$.
In order to capture the combinatorial atom-and-bond structures of molecular graphs,
we decompose the $P_{\mathcal{M}} (M)$ into two parts:
\begin{equation}\small
P_{\mathcal{M}}(M)=P_{\mathcal{M}}((A,B)) \approx P_{\mathcal{A|B}}(A|B; \theta_\mathcal{A|B}) P_{\mathcal{B}}(B;\theta_\mathcal{B}) 
\end{equation}
where   
$P_{\mathcal{M}}$ is the distribution of molecular graphs, 
$P_{\mathcal{B}}$ is the distribution  of bonds (edges) in analogy to modelling multi-channel images , and 
$P_{\mathcal{A|B}}$ is the conditional distribution of atoms (nodes) given the bonds by leveraging graph convolution operations.  The $\theta_\mathcal{B}$ and $\theta_\mathcal{A|B}$ are learnable modelling parameters. 
In contrast with VAE or GAN based frameworks, we can learn the parameters by exact maximum likelihood estimation (MLE) framework by maximizing:
\begin{equation}\small
\argmax_{\theta_\mathcal{B}, \theta_\mathcal{A|B}}  \ \mathbb{E}_{M=(A,B) \sim p_{\mathcal{M}-data}} [\log P_{\mathcal{A|B}}(A|B; \theta_\mathcal{A|B}) + \log P_{\mathcal{B}}(B;\theta_\mathcal{B})]
\end{equation}
Our model thus consists of two parts, namely a \emph{graph conditional  flow for atoms} to learn the atom matrix conditional on the   bond tensors and 
a \emph{flow for bonds} to learn  bond tensors. We further learn a mapping between the learned latent vectors and molecular properties to regress the graph-based molecular  properties, and to guide the generation of optimized molecular graphs.

\subsection{Graph Conditional Flow for Atoms}
\label{sec:model:gcf}
Given a bond tensor $B \in \mathcal{B} \subset \mathbb{R}^{c \times n \times n}$, our goal of the atom flow is to generate the right atom-type matrix $A \in \mathcal{A} \subset \mathbb{R}^{n \times k} $ to assemble valid molecules $M=(A,B) \in \mathcal{M} \subset \mathbb{R}^{n \times k + c \times n \times n} $. We first define \mytag{$\mathcal{B} $-conditional flow} and \mytag{graph conditional flow} $f_{\mathcal{A|B}}$ to transform $A$ given $B$ into conditional latent variable ${Z_{A|B}} = f_{\mathcal{A|B}} (A|B)$ which follows isotropic Gaussian $P_{\mathcal{{Z_{A|B}}}}$.
We can get the conditional probability of atom features given the bond graphs $P_{\mathcal{A|B}}$ by a conditional version of the change of variable formula.

\subsubsection{$\mathcal{B} $-Conditional Flow and Graph Conditional Flow}
\begin{definition} {\mytag{$\mathcal{B} $-conditional flow}:}
A \mytag{$\mathcal{B} $-conditional flow} \newline $Z_{A|B} | B = f_{\mathcal{A|B}}(A | B)$ is an invertible and dimension-kept  mapping 
and there exists reverse transformation $f_{\mathcal{A|B}}^{-1}(Z_{A|B}|B)=A|B$ where $f_{\mathcal{A|B}} \text{ and } f_{\mathcal{A|B}}^{-1}: \mathcal{A} \times \mathcal{B} \mapsto \mathcal{A}\times \mathcal{B}$.
\end{definition}

The condition $B \in \mathcal{B}$ keeps fixed during the transformation. Under the independent assumption of $\mathcal{A}$ and $\mathcal{B}$, the Jacobian of $f_{\mathcal{A|B}}$ is:
\begin{equation}\small
\begin{aligned}
 \frac{\partial f_{\mathcal{A|B}}}{\partial (A,B)} = \begin{bmatrix}
\frac{\partial f_{\mathcal{A|B}}}{\partial A} & \frac{\partial f_{\mathcal{A|B}}}{\partial B}\\
0 & \mathbb{1}_B
\end{bmatrix},
\end{aligned}
\end{equation}
the determiant of this Jacobian is $\det \frac{\partial f_{\mathcal{A|B}}}{\partial (A,B)} = \det \frac{\partial f_{\mathcal{A|B}}}{\partial A}$, and thus the  \textit{conditional version of the change of variable formula} in the form of log-likelihood is: 
\begin{equation} \small
\begin{aligned}
\log P_\mathcal{A|B} (A | B) = \log P_{\mathcal{Z_{A|B}}}(Z_{A|B}) + \log \mid \det \frac{\partial f_{\mathcal{A|B}}}{\partial A} \mid.
\end{aligned}
\end{equation} 

\begin{definition} {\mytag{Graph conditional flow}:}
A graph conditional flow is a $\mathcal{B} $-conditional flow $Z_{A|B} | B= f_{\mathcal{A|B}}(A|B)$  where  $B \in \mathcal{B} \subset \mathbb{R}^{c \times n \times n}$ is the adjacency tenor for edges with $c$ types and $A \in \mathcal{A} \subset \mathbb{R}^{n \times k}$ is the feature matrix of the corresponding $n$ nodes.
\end{definition}

\begin{figure}[!tb]
\vspace{-0.2in}
\centering
\includegraphics[width=0.35\textwidth]{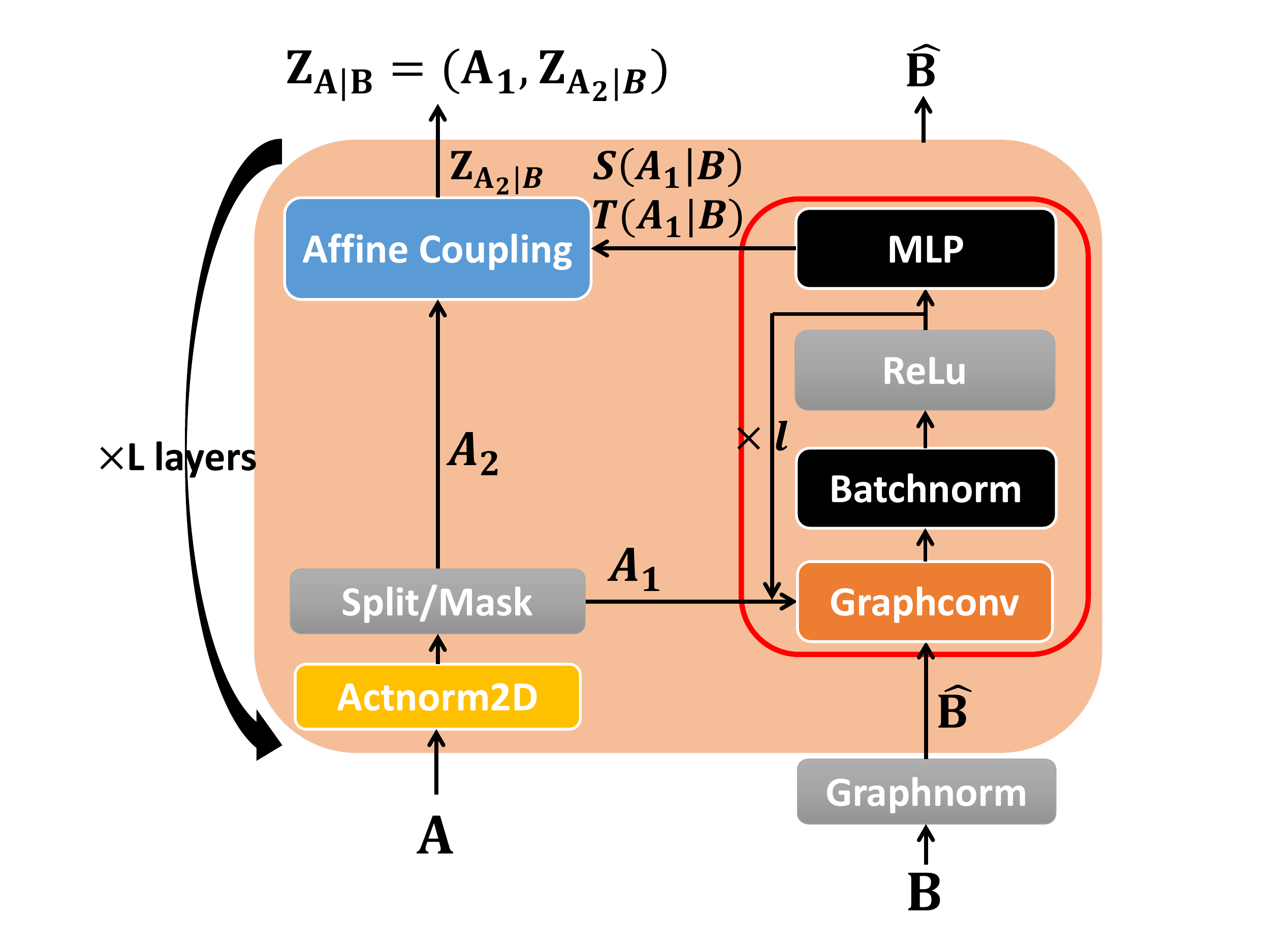}
\vspace{-0.15in}
\caption{ 
Graph conditional flow $f_{\mathcal{A|B}}$ for the atom matrix. We show the details of one invertible graph coupling layer and a multiscale structure consists of a cascade of $L$ layers of such graph coupling layer. The graphnorm is computed only once.
\label{fig:cgflow}}
\vspace{-0.15in}
\end{figure}
\subsubsection{Graph coupling layer} We construct aforementioned invertible mapping $f_{\mathcal{A|B}}$ and $f_{\mathcal{A|B}}^{-1}$ by the  scheme of the affine coupling layer. Different from traditional affine coupling layer, our coupling transformation relies on graph convolution \cite{sun2019graph} and thus we name such a coupling transformation as a \textbf{graph coupling layer}.  

For each graph coupling layer,  we split input $A \in \mathbb{R}^{n \times k}$ into two parts $A=(A_1, A_2)$ along the $n$ row  dimension, and  we get the output $Z_{A|B} = (Z_{A_1|B}, Z_{A_2|B}) = f_{\mathcal{A|B}}(A|B)$ as follows:
\begin{equation}\small
\begin{aligned}
Z_{A_1|B} &= A_{1} \\
Z_{A_2|B} &= A_{2} \odot \text{Sigmoid}(S_{\Theta}(A_{1}| B)) + T_{\Theta}(A_{1}| B)
\end{aligned}
\end{equation} 
where $\odot$ is the element-wise product. We deign the scale function $S_{\Theta}$ and  the transformation function $T_{\Theta}$ in each graph coupling layer by incorporating  \textbf{graph convolution} structures.  The bond tensor $B \in \mathbb{R}^{c \times n \times n}$ keeps a fixed value during  transforming the atom matrix $A$. We also apply the masked convolution idea in \cite{dinh2016density} to the graph convolution in the graph coupling layer.
Here, we adopt Relational Graph Convolutional Networks (R-GCN) \cite{schlichtkrull2018modeling} to build {graph convolution layer} \textbf{graphconv} as follows:
\begin{equation}\small
\begin{aligned}
\text{graphconv}(A_1) = \sum_{i=1}^{c} \hat{B_i}(M\odot A)W_i + (M \odot A)W_0
\end{aligned}
\end{equation} 
where $\hat{B_i} = D^{-1}B_i$ is the normalized adjacency matrix at channel $i$, $D = \sum_{c,i}B_{c,i,j}$ is the sum of the in-degree over all the channels for each node, and $M \in \{0,1\}^{n \times k}$ is a binary mask to select a partition $A_1$ from A. Because the bond graph is fixed during graph coupling layer and thus the graph normalization, denoted as \textbf{graphnorm}, is  computed only once.

We use multiple stacked graphconv->BatchNorm1d->ReLu layers with a multi-layer perceptron (MLP) output layer to build the graph scale function $S_{\Theta}$ and the graph transformation function $T_{\Theta}$. What's more, instead of using exponential function  for the $S_{\Theta}$ as discussed in Sec.~\ref{sec:mp}, we adopt Sigmoid function for the sake of the numerical stability of cascading multiple flow layers.
The reverse mapping of the graph coupling layer $f_{\mathcal{A|B}}^{-1}$ is:
\begin{equation}
\small
\begin{aligned}
A_{1} &= Z_{A_1|B} \\
A_{2} &= (Z_{A_2|B} - T_{\Theta}(Z_{A_1|B}| B))/ \text{Sigmoid}(S_{\Theta}(Z_{A_1|B} | B)).
\end{aligned}
\end{equation} 
The logarithm of the Jacobian determiant of each graph coupling layer can be efficiently computed by:
\begin{equation}
\small
\begin{aligned}
 \log \mid \det (\frac{\partial f_{\mathcal{A|B}}}{\partial A}) \mid=\sum_j \log \text{Sigmoid}(S_{\Theta}(A_1 | B))_j
 \end{aligned}
\end{equation} 
where $j$ iterates each element. In principle, we can use arbitrary complex graph convolution structures for $S_{\Theta}$ and $T_{\Theta}$ since the computing of above Jacobian determinant of $f_{\mathcal{A|B}}$ does not involve in computing the Jacobian of $S_{\Theta}$ or $T_{\Theta}$.

\subsubsection{Actnorm for 2-dimensional matrix} 
For the sake of numerical stability, we design a variant of invertible actnorm layer \cite{kingma2018glow} for the 2-dimensional atom matrix, denoted as \textbf{actnorm2D} (activation normalization for 2D matrix), to normalize each row, namely the feature dimension for each node, over a batch of 2-dimensional atom matrices.  Given the mean $\mu \in \mathbb{R}^{n \times 1}$ and the standard deviation $\sigma^2 \in \mathbb{R}^{n \times 1}$ for each row dimension, the normalized input follows $\hat{A} = \frac{A-\mu}{\sqrt{\sigma^2 + \epsilon}}$ where $\epsilon$ is a small constant, the reverse transformation is $A = \hat{A} * \sqrt{\sigma^2 + \epsilon} + \mu$, and the logarithmic Jacobian determiant is:
\begin{equation}
\small
\begin{aligned}
\log \mid \det \frac{\partial \textbf{actnorm2D}}{\partial  X} \mid = \frac{k}{2}\sum_i^{n}\mid \log({\sigma^2_i + \epsilon}) \mid
\end{aligned}
\end{equation}

\subsubsection{Deep architectures}
We summarize our deep graph conditional  flow in Figure~\ref{fig:cgflow}. We stack multiple graph coupling layers to form graph conditional flow.
We alternate different partition of $A=(A_1, A_2)$ in each layer to transform the unchanged part of the previous layer.

\subsection{Glow for Bonds}
\label{sec:model:glow}
The bond flow aims to learn an invertible mapping $f_{\mathcal{B}}: \mathcal{B} \subset \mathbb{R}^{c \times n \times n} \mapsto \mathcal{B} \subset \mathbb{R}^{c \times n \times n}$ where the transformed latent variable $Z_B = f_{\mathcal{B}} (B)$ follows isotropic Gaussian. According to the change of variable formula, we can get the logarithmic probability of bonds by $\log P_{\mathcal{B}}(B)=\log P_{\mathcal{Z_B}}(Z_B) + \log \mid \det (\frac{\partial f_{\mathcal{B}}}{\partial B}) \mid$ and generating bond tensor by reversing the mapping $\tilde{B} = f_{\mathcal{B}}^{-1} (\tilde{Z})$ where $\tilde{Z} \sim P_{\mathcal{Z}}(Z)$. We can use arbitrary flow model for the bond tensor and we build our bond flow  $f_{\mathcal{B}}$ based on a variant of Glow \cite{kingma2018glow} framework. 

We also follow the scheme of affine coupling layer to build invertible mappings. 
For each affine coupling layer,  We split input $B \in \mathbb{R}^{c \times n \times n}$ into two parts $B=(B_1, B_2)$ along the channel $c$ dimension, and  we get the output $Z_B=(Z_{B_1}, Z_{B_2})$ as follows:
\begin{equation}
\small
\begin{aligned}
Z_{B_1} &= B_{1} \\
Z_{B_2} &= B_{2} \odot \text{Sigmoid}(S_{\Theta}(B_{1})) + T_{\Theta}(B_{1}) .
\end{aligned}
\end{equation} 
And thus the reverse mapping $f_{\mathcal{B}}^{-1}$ is:
\begin{equation}
\small
\begin{aligned}
B_{1} &= Z_{B_1} \\
B_{2} &= (Z_{B_2} - T_{\Theta}(Z_{B_1}))/ \text{Sigmoid}(S_{\Theta}(Z_{B_1})).
\end{aligned}
\end{equation} 
Instead of using exponential function as scale function, we use the Sigmoid function with range $(0,1)$ to ensure the numerical stability when stacking many layers. We find that exponential scale function leads to a large reconstruction error when the number of affine coupling layers increases. The scale function $S_{\Theta}$ and the transformation function $T_{\Theta}$ in each affine coupling layer can have arbitrary structures.  We use multiple $3\times3$ conv2d->BatchNorm2d->ReLu layers to build them. The logarithm of the Jacobian determiant of each affine coupling is 
\begin{equation}
\small
\begin{aligned}\log \mid \det (\frac{\partial Z_B}{\partial B}) \mid=\sum_j \log \text{Sigmoid}(S_{\Theta}(B_1))_j.
\end{aligned}
\end{equation}

\begin{figure}[!tb]
\vspace{-0.3in}
\centering
\includegraphics[width=0.35\textwidth]{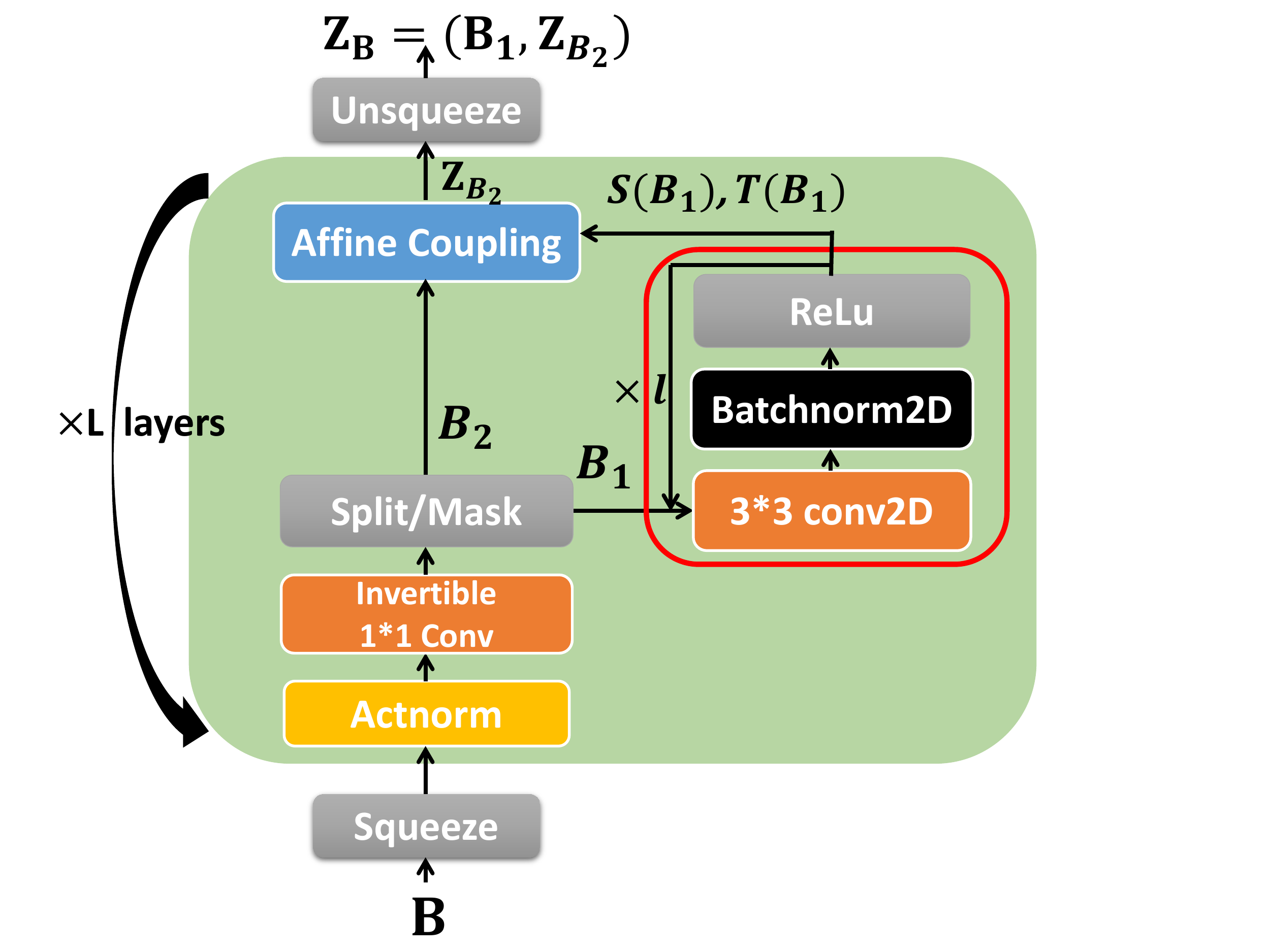}
\label{fig:syn_exp1}
\vspace{-0.15in}
\caption{ 
A variant of Glow  $f_{\mathcal{B}}$ for  bonds' adjacency tensors.
\label{fig:glow}}
\vspace{-0.3in}
\end{figure}

In order to learn optimal partition and ensure model's stability and learning rate, we also use the 
invertible $1 \times 1$ convolution layer and 
actnorm layer adopted in the Glow.
In order to get more channels for masking and transformation, we \mytag{squeeze} the spatial size of $B$ from $\mathbb{R}^{c \times n \times n}$ to $\mathbb{R}^{(c*h*h) \times \frac{n}{h} \times \frac{n}{h}}$  by a factor $h$ and apply the affine coupling transformation to the squeezed data. The reverse \mytag{unsqueeze} operation is adopted to the output.
We summarize our bond flow in Figure~\ref{fig:glow}.

\subsection{Validity Correction}
\label{sec:model:validity}
 Molecules must follow the valency constraints for each atom, but assembling a molecule from generated bond tensor and atom matrix may lead to chemically invalid ones. Here we define the valency constraint for the $i^{th}$ atom as:
 \begin{equation}
 \small
\begin{aligned}
\sum_{c,j} c\times B(c,i,j) \le \text{Valency}(\text{Atom}_i) + Ch
\end{aligned}
\end{equation}
where $B \in \{0,1\}^{c\times n \times n}$ is the one-hot bond tensor over $c \in \{1,2,3\}$ order of chemical bonds (single, double, triple)  and $Ch \in \mathbb{N}$ represents the formal charge. Different from existing valency constraints defined in \cite{you2018graph, popova2019molecularrnn}, we consider the effect of formal charge which may introduce extra bonds for the charged atoms. For example, ammonium [NH4]$^+$ may have 4 bonds for N instead of 3. Similarly, S$^+$ and O$^+$ may have 3 bonds instead of 2.  Here we only consider $Ch=1$ for N$^+$, S$^+$ and O$^+$ and make $Ch=0$ for other atoms.

In contrast with the existing reject-sampling-based validity check adopted in the autoregressive models \cite{you2018graph, popova2019molecularrnn}, we introduce a new post-hoc validity correction procedure 
after generating a molecule $M$ at once: 1) check the valency constraints of $M$; 2) if all the atoms of $M$ follows valecny constraints, we return the largest connected component of the molecule $M$ and end the procedure; 3) if there exists an invalid atom $i$, namely $\sum_{c,j} c \times B(c,i,j) > \text{Valency}(\text{Atom}_i) + Ch$, we sort the bonds of $i$ by their order and delete $1$ order for the bond with the largest order; 4) go to step 1). Our validity correction procedure tries to make a minimum change to the existing molecule and to keep the largest connected component as large as possible.



\subsection{Inference and Generation}
\label{sec:model:all}
We summarize the inference (encoding) and generation (decoding)  of molecular graphs by our \model in Algorithm~\ref{alg:inference} and Algorithm~\ref{alg:generate} respectively. We visualize the overall framework in Figure~\ref{fig:model}. As shown in the algorithms, our \model have merits of exact likelihood estimation/training, one-pass inference, invertible and one-pass generation, and chemical validity guarantee.
\begin{algorithm}[!h]
\footnotesize
\SetAlgoLined
 \textbf{Input:}  $f_\mathcal{A|B}$: graph conditional flow  for atoms, $f_\mathcal{B}$: glow  for bonds, $A$: atom matrix, $B$: bond tensor, $P_{\mathcal{Z_*}}$: isotropic Gaussian distributions.\\
\textbf{Output:} 
$Z_M$:latent representation for atom $M$, 
$\log P_\mathcal{M}(M)$: logarithmic  likelihood of molecule $M$. \\
\quad    $Z_B  = f_\mathcal{B}(B)$ \\
\quad    $\log P_{\mathcal{B}}(B)=\log P_{\mathcal{Z_B}}(Z_B) + \log \mid \det (\frac{\partial f_{\mathcal{B}}}{\partial B}) \mid$ \\
\quad   $\hat{B} = \text{graphnorm(B)}$ \\
\quad   $Z_{A|B} = f_\mathcal{A|B}(A|\hat{B})$ \\
\quad    $ \log P_\mathcal{A|B} (A | B) = \log P_{\mathcal{Z_{A|B}}}(Z_{A|B}) + \log \mid \det (\frac{\partial f_{\mathcal{A|B}}}{\partial A} )\mid $ \\
\quad $Z_M = (Z_{A|B}, Z_B)$\\
\quad   $\log P_{\mathcal{M}} (M) = \log P_\mathcal{B}(B) + \log P_\mathcal{A|B}(A|B)$ \\
\quad \textbf{Return:} $Z_M$, $\log P_\mathcal{M}(M)$\\
 \caption{Exact Likelihood Inference (Encoding) of Molecular Graphs by \model \label{alg:inference}}
\end{algorithm}

\begin{algorithm}[!h]
\footnotesize
\SetAlgoLined
 \textbf{Input:} $f_\mathcal{A|B}$: graph conditional flow  for atoms, $f_\mathcal{B}$: glow  for bonds,  
 $Z_M$:latent representation of molecule $M$ or sampling from a prior Gaussian, 
 validity-correction:  validity correction rules.\\
\textbf{Output:}  $M$: a molecule\\
\quad    $(Z_{A|B}, Z_B) = Z_M$\\
\quad    $B = f_\mathcal{B}^{-1}(Z_B)$ \\
\quad   $\hat{B} = \text{graphnorm(B)}$ \\
\quad   $A  = f_\mathcal{A|B}^{-1}(Z_{A|B}|\hat{B})$ \\
\quad   $M = \text{validity-correction}(A,B)$\\
\quad \textbf{Return:} $M$\\
 \caption{Molecular Graph Generation (Decoding)  by  the Reverse Transformation of \model \label{alg:generate}}
\end{algorithm}

\section{Experiments}
\label{sec:exp}

Following previous works \cite{jin2018junction,shi2020graphaf}, we validate our \model by answering following questions:
\bit
\item \mytag{Molecular graph generation and reconstruction (Sec.~\ref{sec:exp:generation}):}
Can our \model memorize and reconstruct all the training molecule datasets? Can our \model generalize to generate novel, unique and valid molecules as many as possible?

\item \mytag{Visualizing continuous latent space (Sec.~\ref{sec:exp:viz}):}
Can our \model embed molecular graphs into continuous latent space with reasonable chemical similarity?

\item \mytag{Property optimization (Sec.~\ref{sec:exp:opt}):}
Can our \model generate novel molecular graphs with optimized properties?

\item \mytag{Constrained property optimization (Sec.~\ref{sec:exp:copt}):}
Can our \model generate novel molecular graphs with the optimized properties and at the same time keep the chemical similarity as much as possible?
\eit

\mytag{Baselines.} 
We compare our \model with: a) the state-of-the-art VAE-based method JT-VAE \cite{jin2018junction} which captures the chemical validity by encoding and decoding a tree-structured scaffold of molecular graphs; b) the state-of-the-art autoregressive models GCPN \cite{you2018graph} and MolecularRNN (MRNN)\cite{popova2019molecularrnn} with reinforcement learning for property optimization, which generate molecules in a sequential manner; c) flow-based methods GraphNVP \cite{madhawa2019graphnvp} and GRF \cite{honda2019graph} which generate molecules at one shot and the state-of-the-art autoregressive-flow-based model GraphAF \cite{shi2020graphaf} which generates molecules in a sequential way.

\mytag{Datasets.} 
We use two datasets QM9 \cite{ramakrishnan2014quantum} and ZINC250K \cite{irwin2012zinc} for our experiments and summarize them in Table~\ref{tab:data}. The QM9 contains $133,885$ molecules with maximum $9$ atoms in $4$ different types, and the ZINC250K has $249,455$ drug-like molecules with maximum $38$ atoms in $9$ different types. The molecules are kekulized by the chemical software RDKit \cite{landrum2006rdkit} and the hydrogen atoms are removed. There are three types of edges, namely single, double, and triple bonds, for all  molecules. Following the pre-processing procedure in \cite{madhawa2019graphnvp},  we encode each atom and bond by one-hot encoding, pad the molecules which have less than the maximum number of atoms with an virtual atom, augment the adjacency tensor of each molecule by a virtual edge channel representing no bonds between atoms, and dequantize \cite{madhawa2019graphnvp,dinh2016density} the discrete one-hot-encoded data by adding uniform random noise $U[0,0.6]$ for each dimension,
leading to atom matrix $A\in \mathbb{R}^{9\times5}$  and bond tensor $B\in \mathbb{R}^{4 \times 9\times 9}$ for QM9, and $A\in \mathbb{R}^{38\times10}$ and
$B\in \mathbb{R}^{4 \times 38\times 38}$ for ZINC250k. 
\begin{table}[!htb] 
\vspace{-0.15in}
\small
\centering 
\caption{Statistics of the datasets.}
\vspace{-0.05in}
\begin{tabular}
{ l p{1.2cm} p{1.2 cm} p{1.2cm} p{1.2cm} p{1.2cm}  }
\toprule 
 \bf{} &  \bf{\#Mol. Graphs} &  \bf{Max. \#Nodes} &  \bf{\#Node Types} &  \bf{\#Edge Types}\\ 
\midrule 
QM9 & 133,885 & 9  &  4+1 & 3+1 \\ 
ZINC250K & 249,455 & 38  & 9+1 & 3+1 \\ 
\bottomrule 
\end{tabular}
\label{tab:data} 
\vspace{-0.1in}
\end{table}

\mytag{\modelTitle Setup.} 
To be comparable with one-shot-flow baseline GraphNVP \cite{madhawa2019graphnvp},  for the ZINC250K, we adopt $10$ coupling layers and $38$ graph coupling layers for the bonds' Glow and the atoms' graph conditional flow respectively.  We use two $3*3$ convolution layers with $512,512$ hidden dimensions in each coupling layer. 
For each graph coupling layer, we set one relational graph convolution layer with $256$ dimensions followed by a two-layer multilayer perceptron with $512,64$ hidden dimensions. 
As for the QM9, we adopt $10$ coupling layers and $27$ graph coupling layers for the bonds' Glow and  the atoms' graph conditional flow respectively.  There are two 3*3 convolution layers with $128,128$ hidden dimensions in each coupling layer, and one graph convolution layer with $64$ dimensions followed by a two-layer multilayer perceptron with $128,64$ hidden dimensions in each graph coupling layer. 
 As for the optimization experiments, we further train a regression model to map the latent embeddings to different  property scalars (discussed in Sec.~\ref{sec:exp:opt} and \ref{sec:exp:copt}) by a multi-layer perceptron with 18-dim linear layer -> ReLu -> 1-dim linear layer structures. For each dataset, we use the same trained model for all the following experiments.

\mytag{Empirical Running Time.} 
Following above setup, we implemented our \model by Pytorch-1.3.1 and trained it by Adam optimizer \cite{kingma2014adam} with learning rate $0.001$, batch size $256$, and $200$ epochs for both datasets on $1$ GeForce RTX 2080 Ti GPU and 16 CPU cores. 
Our \model finished $200$-epoch training within $22$ hours ($6.6$ minutes/epoch) for ZINC250K  and $3.3$ hours ($0.99$ minutes/epoch)  for QM9. Thanks to efficient one-pass inference/embedding, our \model takes negligible $7$ minutes to learn an additional regression layer trained in $3$ epochs for optimization experiments on ZINC250K. In comparison, as for the ZINC250K dataset, GraphNVP \cite{madhawa2019graphnvp} costs $38.4$ hours ($11.5$ minutes/epoch) by our Pytorch implementation for training on ZINC250K with the same configurations, and the estimated total running time of GraphAF \cite{shi2020graphaf} is $124$ hours  ($24$ minutes/epoch) which consists of the reported $4$ hours for a generation model trained by $10$ epochs  and estimated $120$  hours for another optimization model trained by $300$ epochs. The reported running time of JT-VAE \cite{jin2018junction} is roughly $24$ hours in \cite{you2018graph}. 






  
\subsection{Generation and Reconstruction}
\label{sec:exp:generation}
\begin{table*}[!htb] 
\small
\centering 
\caption{Generation and reconstruction performance on QM9 dataset. }
\vspace{-0.1in}
\begin{tabular}{l c c c c c c } 
\toprule 
 & \textbf{\% Validity} & \textbf{\% Validity w/o check} & \textbf{\% Uniqueness} &  \textbf{\% Novelty}   & \textbf{\% N.U.V.}  & \textbf{\% Reconstruct}\\ 
\midrule 
GraphNVP \cite{madhawa2019graphnvp} & $83.1\pm0.5$  & n/a  &  $99.2\pm0.3$& $58.2\pm1.9$    & $47.97$ & $100$\\ 
GRF \cite{honda2019graph}& $84.5\pm 0.70$  & n/a  &  $66.0\pm1.15$& $58.6\pm 0.82$    & $32.68$ &$100$\\ 
GraphAF \cite{shi2020graphaf}& $100$         & $67$& $94.51$ & $88.83$    & $83.95$ &$100$\\ 
\midrule
\textbf{MoFlow} & $\mathbf{100.00\pm0.00}$ &  $\mathbf{96.17\pm0.18}$ &  $\mathbf{99.20\pm0.12}$ & $\mathbf{98.03\pm0.14}$  & $\mathbf{97.24\pm0.21}$ & $\mathbf{100.00\pm0.00}$ \\ 
\bottomrule 
\end{tabular}
\label{tab:qm9} 
\end{table*}

\begin{table*}[!htb] 
\small
\centering 
\caption{Generation and reconstruction performance on ZINC250K dataset.
}
\vspace{-0.1in}
\begin{tabular}{l c c c c c c } 
\toprule 
 & \textbf{\% Validity} & \textbf{\% Validity w/o check} & \textbf{\% Uniqueness} &  \textbf{\% Novelty}   & \textbf{\% N.U.V.}  & \textbf{\% Reconstruct}\\ 
\midrule 
JT-VAE \cite{jin2018junction} & $100$  & n/a  &  $100$& $100$    & $100$ & $76.7$\\ 
GCPN \cite{you2018graph} & $100$  & $20$  &  $99.97$& $100$    & $99.97$ & n/a\\ 
MRNN \cite{popova2019molecularrnn}& $100$  & $65$  &  $99.89$ & $100$    & $99.89$ & n/a\\ 
GraphNVP \cite{madhawa2019graphnvp} & $42.6\pm1.6$  & n/a  &  $94.8\pm0.6$& $100$    & $40.38$ & $100$\\ 
GRF \cite{honda2019graph} & $73.4\pm 0.62$  & n/a  &  $53.7\pm 2.13$& $100$    & $39.42$ &$100$\\ 
GraphAF \cite{shi2020graphaf}& $100$         & $68$& $99.10$ & $100$    & $99.10$ &$100$\\ 
\midrule
\textbf{MoFlow} & $\mathbf{100.00\pm0.00}$ &  $\mathbf{81.76\pm0.21}$ &  $\mathbf{99.99\pm0.01}$ & $\mathbf{100.00\pm0.00}$  & $\mathbf{99.99\pm0.01}$ & $\mathbf{100.00\pm0.00}$ \\ 
\bottomrule 
\end{tabular}
\label{tab:zinc} 
\end{table*}
\mytag{Setup.} In this task, we evaluate our \model's capability of generating novel, unique and valid molecular graphs, and if our \model can reconstruct input molecular graphs from their latent representations.  We adopted the widely-used metrics, including: \textbf{Validity} which is the percentage of chemically valid molecules in all the generated molecules, \textbf{Uniqueness} which is the percentage of unique valid molecules in all the generated molecules, \textbf{Novelty} which is the percentage of generated valid molecules which are not in the training dataset, and \textbf{Reconstruction} rate which is the percentage of molecules in the input dataset which can be reconstructed from their latent representations. Besides, because the novelty score also accounts for the potentially duplicated novel molecules, we propose a new metric \textbf{N.U.V.} which is the percentage of \underline{N}ovel, \underline{U}nique, and \underline{V}alid molecules in all the generated molecules. We also compare the validity of ablation models if not using validity check or validity correction, denoted as \textbf{Validity w/o check} in \cite{shi2020graphaf}. 

The prior distribution of latent space follows a spherical multivariate Gaussian distribution
$\mathcal{N}(0, {(t \sigma)}^2 \mathbf{I})$ where $\sigma$ is the learned standard deviation and the hyper-parameter $t$ is the temperature for the reduced-temperature generative model \cite{parmar2018image, kingma2018glow,madhawa2019graphnvp}. We use $t=0.85$ in the generation for both QM9 and ZINC250K datasets, and $t=0.6$ for the ablation study without validity correction. To be comparable with the state-of-the-art baseline GraphAF\cite{shi2020graphaf}, we generate $10,000$ molecules, i.e., sampling  $10,000$ latent vectors from the prior and then decode them by the reverse transformation of our \modelEOL.   We report the the mean and standard deviation of results over 5 runs. 
As for the reconstruction,  we encode all the molecules from the training dataset into latent vectors by the encoding transformation of our \model and then reconstruct input molecules from these latent vectors  by the reverse transformation of \modelEOL.

\mytag{Results.} Table~\ref{tab:qm9} and Table~\ref{tab:zinc} show that our \model outperfoms the state-of-the-art models on all the six metrics for both QM9 and ZINC250k datasets. 
Thanks to the invertible  characteristic of the flow-based models, our \model builds an one-to-one mapping from the input molecule $M$ to its corresponding latent vector $Z$, enabling $100\%$ reconstruction rate as shown in Table~\ref{tab:qm9} and Table~\ref{tab:zinc}. In contrast, the VAE-based method JT-VAE and the autoregressive-based method GCPN and MRNN can't  reconstruct all the input molecules. Compared with the one-shot flow-based model GraphNVP and GRF, by incorporating validity correction mechanism, our \model achieves $100\%$ validity, leading to significant improvements of the validity score and N.U.V. score for both datasets. Specifically, the N.U.V. score of \model are $2$ and $3$ times as large as the N.U.V. scores of GraphNVP and GRF respectively in Table~\ref{tab:qm9}. 
Even without validity correction, our \model still outperforms 
the validity scores of GraphNVP and GRF by a large margin.
Compared with the autoregressive flow-based model GraphAF,  we find 
our \model outperforms GraphAF
by additional $16\%$ and $0.8\%$ with respect to N.U.V scores for QM9 and ZINC respectively, indicating that our \model generates more novel, unique and valid molecules. Indeed, \model achieves better uniqueness score and novelty score compared with GraphAF for both datasets. What's more, our \model without  validity correction still outperforms GraphAF without the validity check by a large margin w.r.t. the validity score (validity w/o check in Table~\ref{tab:qm9} and Table~\ref{tab:zinc}) for both datasets, implying the superiority of capturing the molecular structures in a holistic way by our \model over  autoregressive ones in a sequential way.

In conclusion, our \model not only memorizes and reconstructs all the training molecular graphs, but also generates more novel, unique and valid molecular graphs than existing models, indicating that our \model  learns a strict superset of the training data and explores the unknown chemical space better.


\subsection{Visualizing Continuous Latent Space}
\label{sec:exp:viz}
\begin{figure*}[!th]
\vspace{-0.2in}
\centering
\includegraphics[width=.6\textwidth]{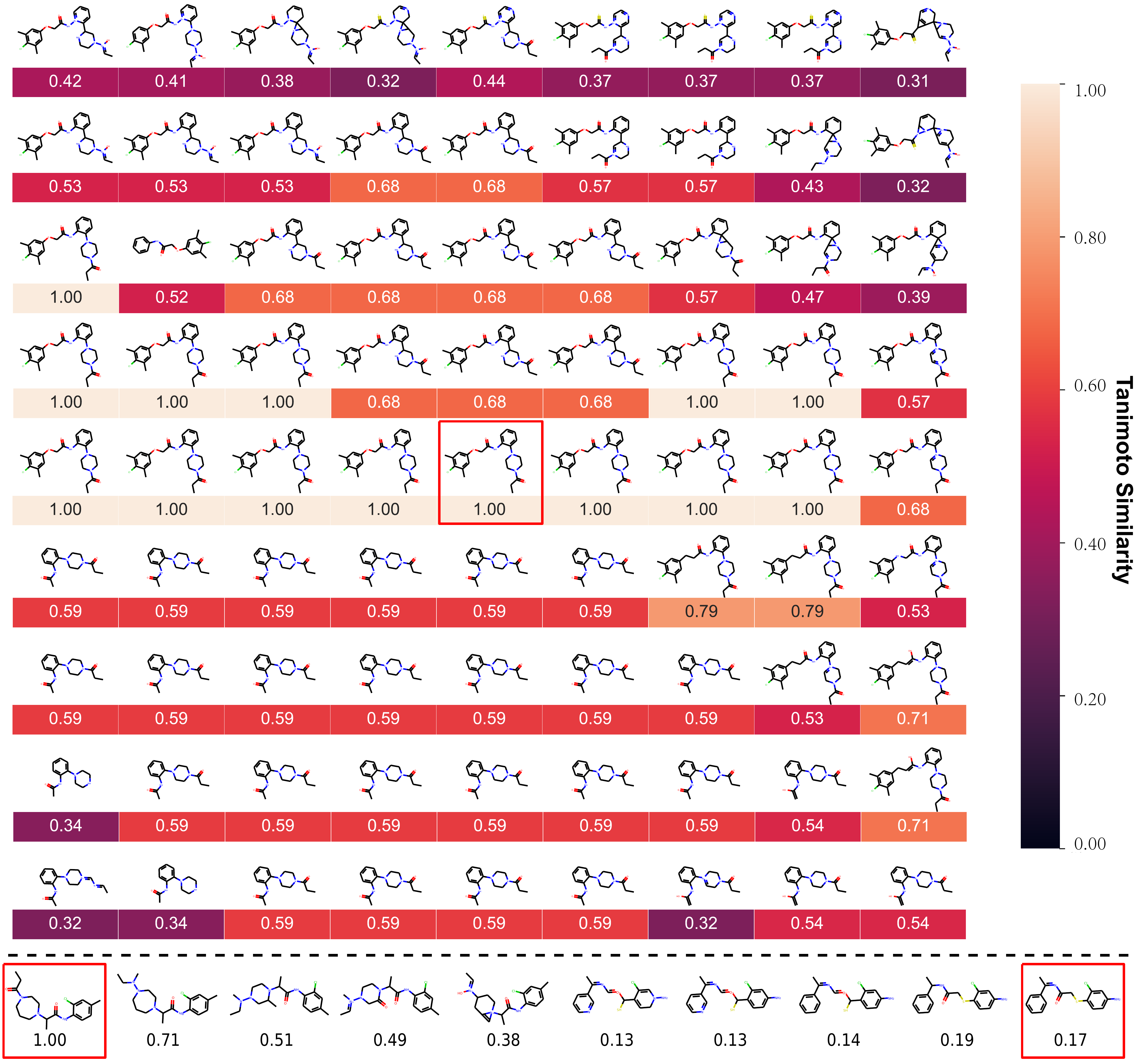}
\vspace{-0.1in}
\caption{ 
Visualization of learned latent space by our \modelEOL. Top: Visualization of the grid neighbors of a seed molecule in the center, which serves as the baseline for measuring similarity. Bottom: Interpolation between two seed molecular graphs and the left one is the baseline molecule for measuring similarity. Seed molecules are highlighted in red boxs and they are randomly selected from ZINC250K. 
\label{fig:zincint}}
\end{figure*}


\mytag{Setup.} We examine the learned latent space of our \model, denoted as $f$, by visualizing the decoded molecular graphs from a neighborhood of a latent vector in the latent space. Similar to \cite{kusner2017grammar, jin2018junction}, we encode a seed molecule $M$ into $Z=f(M)$ and then grid search two random orthogonal directions with unit vector $X$ and $Y$ based on $Z$, then we get new latent vector by $Z' = Z + \lambda_X*X + \lambda_Y*Y$ where $\lambda_X$ and $\lambda_Y$ are the searching steps. Different from VAE-based models, our \model  gets decoded molecules efficiently by the one-pass inverse transformation $M'=f^{-1}(Z')$. In contrast, the VAE-based models such as JT-VAE need to decode each latent vectors $10-100$ times and autoregressive-based models like GCPN, MRNN and GraphAF need to generate a molecule sequentially.
Further more, we measure the chemical similarity between each neighboring molecule and the centering molecule. We choose Tanimoto index \cite{bajusz2015tanimoto} as the chemical similarity metrics and indicate their similarity values by a heatmap. We further visualize a linear interpolation between two molecules to show their changing trajectory 
similar to the interpolation case between images \cite{kingma2018glow}.

\mytag{Results.} We show the visualization of latent space in Figure~\ref{fig:zincint}. We find the latent space is very smooth and the interpolations between two latent points only change a molecule graph a little bit. Quantitatively, we find the chemical similarity between molecules majorly correspond to their Euclidean distance between their latent vectors, implying that our \model embeds similar molecular graph structures into similar latent embeddings.  Searching in such a continuous latent space learnt by our \model is the basis for molecular  property optimization and constraint optimization as discussed in the following sections.




\vspace{-0.1in}
\subsection{Property Optimization}
\label{sec:exp:opt}
\mytag{Setup.} 
The property optimization task aims at generating novel molecules with the best Quantitative Estimate of Druglikeness (QED)  scores \cite{bickerton2012quantifying} which measures the drug-likeness of generated molecules.  Following the previous works \cite{you2018graph,popova2019molecularrnn}, we report the  best  property scores of novel molecules discovered by each method.

We use the pre-trained \modelEOL, denoted as $f$, in the generation experiment to encode a molecule $M$ and get the molecular embedding $Z = f(M)$, and further train a multilayer perceptron to regress the embedding $Z$ of the molecules to their property values $y$. We then search the best molecules by the gradient ascend method, namely $Z' = Z + \lambda * \frac{d y}{ dZ}$ where the $\lambda$ is the length of the search step. We conduct above gradient ascend method by $K$ steps. 
We decode the new embedding $Z'$ in the latent space to the discovered molecule by reverse mapping $M' = f^{-1}(Z')$. The molecule $M'$ is novel if $M'$ doesn't exist in the training dataset. 

\begin{table}[!h] 
\vspace{-0.1in}
\footnotesize
\centering 
\caption{Discovered novel molecules with the best QED scores. Our \model finds more molecules with the best QED scores. More results in Figure~\ref{fig:topqed}.}
\vspace{-0.1in}
\begin{tabular}
{ l p{1.2cm} p{1.2 cm} p{1.2cm} p{1.2cm} p{1.2cm}  }
\toprule 
 \bf{Method} &  \bf{1st} &  \bf{2nd} &  \bf{3rd} &  \bf{4th}\\ 
\midrule 
ZINC (Dataset) & 0.948 & 0.948  &  0.948 & 0.948 \\ 
\midrule
JT-VAE & 0.925 & 0.911  &  0.910 & -\\
GCPN & 0.948 & 0.947  &  0.946 & -\\
MRNN & 0.948 & 0.948  &  0.947 & -\\
GraphAF & 0.948 & 0.948  &  0.947 & 0.946\\
\midrule
\bf{\model} & \bf{0.948} & \bf{0.948}  &  \bf{0.948} & \bf{0.948}\\
\bottomrule 
\end{tabular}
\label{tab:topqed} 
\vspace{-0.2in}
\end{table}
\begin{figure}[!h]
\vspace{-0.25in}
\small
\centering
\includegraphics[width=.4\textwidth]{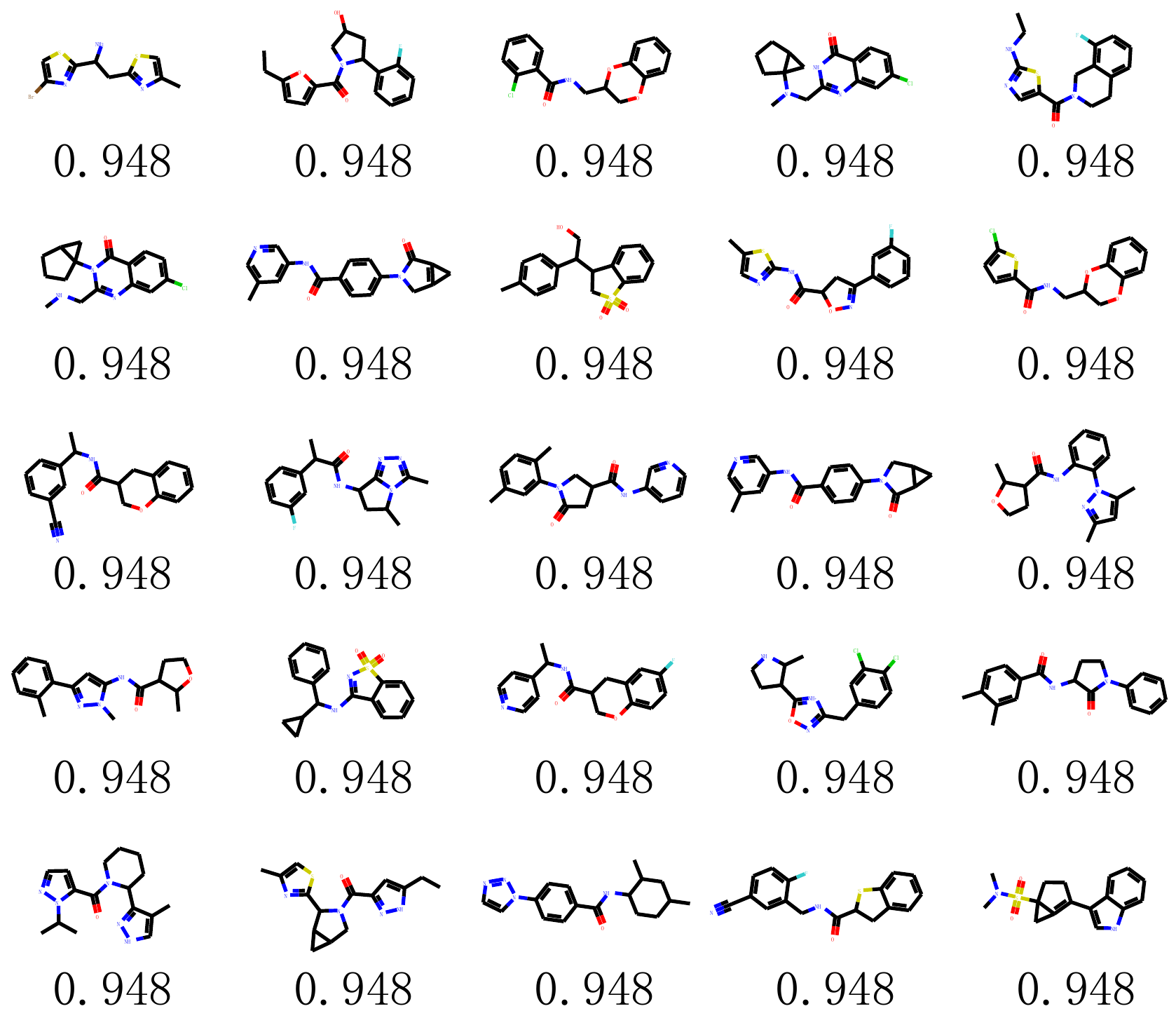}
\label{fig:syn_exp1}
\vspace{-0.1in}
\caption{ 
Illustration of discovered novel molecules with the best druglikeness QED scores.
\label{fig:topqed}}
\vspace{-0.1in}
\end{figure}
\mytag{Results.}
We report the  discovered novel molecules sorted by their QED scores in Table~\ref{tab:topqed}. We find previous methods can only find very few molecules with the best QED score ($=0.948$). In contrast, our \model finds much more novel molecules which have the best QED values than all the  baselines. We show more molecular structures with top QED values in Figure~\ref{fig:topqed}.

\vspace{-0.1in}
\subsection{Constrained Property Optimization}
\label{sec:exp:copt}
\mytag{Setup.}
The constrained property optimization aims at finding a new molecule $M'$ with the largest similarity score $sim(M,M')$ and the largest improvement of a targeted property value $y(M') - y(M)$ given a molecule $M$. Following the similar experimental setup of \cite{jin2018junction,you2018graph}, we choose Tanimoto similarity of Morgan fingerprint \cite{rogers2010extended} as the similarity metrics, the penalized logP (plogp) as the target property, and $M$ from the $800$ molecules with the lowest plogp scores in the training dataset of ZINC250K. 
We use similar gradient ascend method as discussed in the previous subsetion to search for optimized molecules.
An optimization succeeds if we find a novel molecule $M'$ which is different from $M$ and $y(M') - y(M) \ge 0$ and $sim(M,M') \ge \delta$  within $K$ steps where $\delta$ is the smallest similarity threshold to screen the optimized molecules.

\begin{figure}[!t]
\vspace{-0.1in}
\centering
\includegraphics[width=0.4\textwidth]{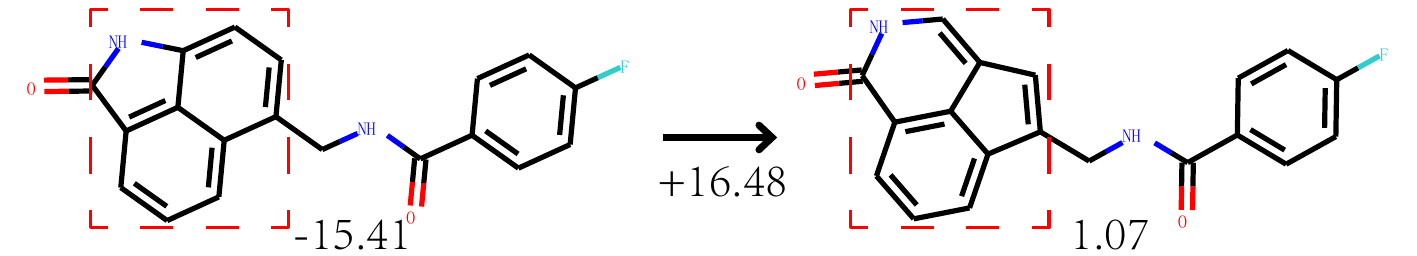}
\vspace{-0.1in}
\caption{ 
An illustration of the constrained optimization of a  molecule leading to an improvement of $+16.48$ w.r.t the penalized logP and with Tanimoto similarity $0.624$. The modified part is highlighted.
\label{fig:copt}}
\vspace{-0.1in}
\end{figure}

\mytag{Results.} Results are summarized in Table~\ref{tab:plogp}. We find that our \model finds the most similar new molecules at the same time achieves very good plogp improvement. Compared with the state-of-the-art VAE model JT-VAE, our \model achieves much higher similarity score and property improvement, implying that our model is good at interpolation and learning continuous molecular embedding. Compared with the state-of-the-art reinforcement learning based method GCPN and GraphAF which is good at generating molecules step-by-step with targeted property rewards, our model \model achieves the best similarity scores and the second best property improvements. We illustrate one optimization example in Figure~\ref{fig:copt} with very similar structures but a large improvement w.r.t the penalized logP.

\begin{table}[!tb] 
\vspace{-0.05in}
\scriptsize
\centering 
\caption{Constrained optimization on Penalized-logP}
\vspace{-0.1in}
\begin{tabular}{l c c c c c c c} 
\toprule 
& \multicolumn{3}{c}{JT-VAE} & \multicolumn{3}{c}{GCPN}\\ 
\cmidrule(l){2-4}   \cmidrule(l){5-7} 
 $\delta$ & \textbf{Improvement} &  \textbf{Similarity} & \textbf{Success} & \textbf{Improvement} &  \textbf{Similarity} & \textbf{Success} \\ 
\midrule 
\textbf{0.0} & $ 1.91\pm 2.04$ & $ 0.28\pm 0.15$  &  $97.5\%$ &$ 4.20\pm 1.28$  & $ \mathbf{0.32\pm 0.12}$& $100\%$\\ 
\textbf{0.2} & $ 1.68\pm 1.85$ & $ 0.33\pm 0.13$  &  $97.1\%$ &$4.12 \pm 1.19$  & $ 0.34 \pm 0.11$& $100\%$\\ 
\textbf{0.4} & $0.84 \pm 1.45$ & $0.51 \pm 0.10$  &  $83.6\%$ &$2.49 \pm 1.30$  & $ 0.48\pm 0.08$& $100\%$\\ 
\textbf{0.6} & $ 0.21\pm 0.71$ & $ 0.69\pm 0.06$  &  $46.4\%$ &$0.79 \pm 0.63$  & $ 0.68\pm 0.08$& $100\%$\\ 
\midrule
& \multicolumn{3}{c}{GraphAF} & \multicolumn{3}{c}{\textbf{\model}}\\ 
\cmidrule(l){2-4}   \cmidrule(l){5-7} 
 $\delta$ & \textbf{Improvement} &  \textbf{Similarity} & \textbf{Success} & \textbf{Improvement} &  \textbf{Similarity} & \textbf{Success} \\ 
\midrule 
\textbf{0.0} & $ 13.13\pm 6.89$ & $0.29 \pm 0.15$  &  $100\%$ &$ 8.61\pm 5.44$  & $0.30 \pm 0.20 $ & $98.88\%$\\ 
\textbf{0.2} & $ 11.90\pm 6.86$ & $ 0.33\pm 0.12$  &  $100\%$ &$7.06 \pm 5.04$  & $\mathbf{0.43 \pm 0.20 }$& $96.75\%$\\ 
\textbf{0.4} & $ 8.21\pm 6.51$ & $0.49 \pm 0.09$  &  $99.88\%$ &$4.71 \pm4.55 $  & $ \mathbf{0.61\pm0.18} $& $85.75\%$\\ 
\textbf{0.6} & $4.98 \pm 6.49$ & $0.66 \pm 0.05$  &  $96.88\%$ &$ 2.10\pm 2.86$  & $ \mathbf{0.79\pm 0.14}$& $58.25\%$\\ 
\bottomrule 
\bottomrule 
\end{tabular}
\label{tab:plogp} 
\end{table}

\section{Conclusion}
\label{sec:conclusion}
In this paper, we propose a novel deep graph generative model   \model  for molecular graph generation. Our \model is  one of the first flow-based models which not only generates molecular graphs at one-shot by invertible mappings but also has a validity guarantee. 
Our \model consists of a variant of Glow model for bonds, a novel graph conditional flow for atoms given bonds, and then combining them with post-hoc validity corrections. 
Our \model achieves state-of-the-art performance on molecular generation, reconstruction and optimization.
For future work, we try to combine the advantages of both sequential generative models and one-shot generative models to generate chemically feasible molecular graphs. Codes and datasets are open-sourced  at \url{https://github.com/calvin-zcx/moflow}.

\small{
\section*{Acknowledgement}
This work is supported by NSF IIS 1716432, 1750326, ONR N00014-18-1-2585, Amazon Web Service (AWS) Machine Learning for Research Award and Google Faculty Research Award.
}

\small{
\bibliographystyle{ACM-Reference-Format}
\bibliography{acmart}


\begin{thebibliography}{35}


\ifx \showCODEN    \undefined \def \showCODEN     #1{\unskip}     \fi
\ifx \showDOI      \undefined \def \showDOI       #1{#1}\fi
\ifx \showISBNx    \undefined \def \showISBNx     #1{\unskip}     \fi
\ifx \showISBNxiii \undefined \def \showISBNxiii  #1{\unskip}     \fi
\ifx \showISSN     \undefined \def \showISSN      #1{\unskip}     \fi
\ifx \showLCCN     \undefined \def \showLCCN      #1{\unskip}     \fi
\ifx \shownote     \undefined \def \shownote      #1{#1}          \fi
\ifx \showarticletitle \undefined \def \showarticletitle #1{#1}   \fi
\ifx \showURL      \undefined \def \showURL       {\relax}        \fi
\providecommand\bibfield[2]{#2}
\providecommand\bibinfo[2]{#2}
\providecommand\natexlab[1]{#1}
\providecommand\showeprint[2][]{arXiv:#2}

\bibitem[\protect\citeauthoryear{Avorn}{Avorn}{2015}]%
        {avorn20152}
\bibfield{author}{\bibinfo{person}{Jerry Avorn}.}
  \bibinfo{year}{2015}\natexlab{}.
\newblock \showarticletitle{The \$2.6 billion pill—methodologic and policy
  considerations}.
\newblock \bibinfo{journal}{\emph{New England Journal of Medicine}}
  \bibinfo{volume}{372}, \bibinfo{number}{20} (\bibinfo{year}{2015}),
  \bibinfo{pages}{1877--1879}.
\newblock


\bibitem[\protect\citeauthoryear{Bajusz, R{\'a}cz, and H{\'e}berger}{Bajusz
  et~al\mbox{.}}{2015}]%
        {bajusz2015tanimoto}
\bibfield{author}{\bibinfo{person}{D{\'a}vid Bajusz}, \bibinfo{person}{Anita
  R{\'a}cz}, {and} \bibinfo{person}{K{\'a}roly H{\'e}berger}.}
  \bibinfo{year}{2015}\natexlab{}.
\newblock \showarticletitle{Why is Tanimoto index an appropriate choice for
  fingerprint-based similarity calculations?}
\newblock \bibinfo{journal}{\emph{Journal of cheminformatics}}
  \bibinfo{volume}{7}, \bibinfo{number}{1} (\bibinfo{year}{2015}),
  \bibinfo{pages}{20}.
\newblock


\bibitem[\protect\citeauthoryear{Bickerton, Paolini, Besnard, Muresan, and
  Hopkins}{Bickerton et~al\mbox{.}}{2012}]%
        {bickerton2012quantifying}
\bibfield{author}{\bibinfo{person}{G~Richard Bickerton},
  \bibinfo{person}{Gaia~V Paolini}, \bibinfo{person}{J{\'e}r{\'e}my Besnard},
  \bibinfo{person}{Sorel Muresan}, {and} \bibinfo{person}{Andrew~L Hopkins}.}
  \bibinfo{year}{2012}\natexlab{}.
\newblock \showarticletitle{Quantifying the chemical beauty of drugs}.
\newblock \bibinfo{journal}{\emph{Nature chemistry}} \bibinfo{volume}{4},
  \bibinfo{number}{2} (\bibinfo{year}{2012}), \bibinfo{pages}{90}.
\newblock


\bibitem[\protect\citeauthoryear{Bresson and Laurent}{Bresson and
  Laurent}{2019}]%
        {bresson2019two}
\bibfield{author}{\bibinfo{person}{Xavier Bresson} {and}
  \bibinfo{person}{Thomas Laurent}.} \bibinfo{year}{2019}\natexlab{}.
\newblock \showarticletitle{A Two-Step Graph Convolutional Decoder for Molecule
  Generation}.
\newblock \bibinfo{journal}{\emph{arXiv preprint arXiv:1906.03412}}
  (\bibinfo{year}{2019}).
\newblock


\bibitem[\protect\citeauthoryear{Dai, Tian, Dai, Skiena, and Song}{Dai
  et~al\mbox{.}}{2018}]%
        {dai2018syntax}
\bibfield{author}{\bibinfo{person}{Hanjun Dai}, \bibinfo{person}{Yingtao Tian},
  \bibinfo{person}{Bo Dai}, \bibinfo{person}{Steven Skiena}, {and}
  \bibinfo{person}{Le Song}.} \bibinfo{year}{2018}\natexlab{}.
\newblock \showarticletitle{Syntax-directed variational autoencoder for
  structured data}.
\newblock \bibinfo{journal}{\emph{arXiv preprint arXiv:1802.08786}}
  (\bibinfo{year}{2018}).
\newblock


\bibitem[\protect\citeauthoryear{De~Cao and Kipf}{De~Cao and Kipf}{2018}]%
        {de2018molgan}
\bibfield{author}{\bibinfo{person}{Nicola De~Cao} {and} \bibinfo{person}{Thomas
  Kipf}.} \bibinfo{year}{2018}\natexlab{}.
\newblock \showarticletitle{MolGAN: An implicit generative model for small
  molecular graphs}.
\newblock \bibinfo{journal}{\emph{arXiv preprint arXiv:1805.11973}}
  (\bibinfo{year}{2018}).
\newblock


\bibitem[\protect\citeauthoryear{Dinh, Krueger, and Bengio}{Dinh
  et~al\mbox{.}}{2014}]%
        {dinh2014nice}
\bibfield{author}{\bibinfo{person}{Laurent Dinh}, \bibinfo{person}{David
  Krueger}, {and} \bibinfo{person}{Yoshua Bengio}.}
  \bibinfo{year}{2014}\natexlab{}.
\newblock \showarticletitle{Nice: Non-linear independent components
  estimation}.
\newblock \bibinfo{journal}{\emph{arXiv preprint arXiv:1410.8516}}
  (\bibinfo{year}{2014}).
\newblock


\bibitem[\protect\citeauthoryear{Dinh, Sohl-Dickstein, and Bengio}{Dinh
  et~al\mbox{.}}{2016}]%
        {dinh2016density}
\bibfield{author}{\bibinfo{person}{Laurent Dinh}, \bibinfo{person}{Jascha
  Sohl-Dickstein}, {and} \bibinfo{person}{Samy Bengio}.}
  \bibinfo{year}{2016}\natexlab{}.
\newblock \showarticletitle{Density estimation using real nvp}.
\newblock \bibinfo{journal}{\emph{arXiv preprint arXiv:1605.08803}}
  (\bibinfo{year}{2016}).
\newblock


\bibitem[\protect\citeauthoryear{G{\'o}mez-Bombarelli, Wei, Duvenaud,
  Hern{\'a}ndez-Lobato, S{\'a}nchez-Lengeling, Sheberla, Aguilera-Iparraguirre,
  Hirzel, Adams, and Aspuru-Guzik}{G{\'o}mez-Bombarelli et~al\mbox{.}}{2018}]%
        {gomez2018automatic}
\bibfield{author}{\bibinfo{person}{Rafael G{\'o}mez-Bombarelli},
  \bibinfo{person}{Jennifer~N Wei}, \bibinfo{person}{David Duvenaud},
  \bibinfo{person}{Jos{\'e}~Miguel Hern{\'a}ndez-Lobato},
  \bibinfo{person}{Benjam{\'\i}n S{\'a}nchez-Lengeling},
  \bibinfo{person}{Dennis Sheberla}, \bibinfo{person}{Jorge
  Aguilera-Iparraguirre}, \bibinfo{person}{Timothy~D Hirzel},
  \bibinfo{person}{Ryan~P Adams}, {and} \bibinfo{person}{Al{\'a}n
  Aspuru-Guzik}.} \bibinfo{year}{2018}\natexlab{}.
\newblock \showarticletitle{Automatic chemical design using a data-driven
  continuous representation of molecules}.
\newblock \bibinfo{journal}{\emph{ACS central science}} \bibinfo{volume}{4},
  \bibinfo{number}{2} (\bibinfo{year}{2018}), \bibinfo{pages}{268--276}.
\newblock


\bibitem[\protect\citeauthoryear{Honda, Akita, Ishiguro, Nakanishi, and
  Oono}{Honda et~al\mbox{.}}{2019}]%
        {honda2019graph}
\bibfield{author}{\bibinfo{person}{Shion Honda}, \bibinfo{person}{Hirotaka
  Akita}, \bibinfo{person}{Katsuhiko Ishiguro}, \bibinfo{person}{Toshiki
  Nakanishi}, {and} \bibinfo{person}{Kenta Oono}.}
  \bibinfo{year}{2019}\natexlab{}.
\newblock \showarticletitle{Graph residual flow for molecular graph
  generation}.
\newblock \bibinfo{journal}{\emph{arXiv preprint arXiv:1909.13521}}
  (\bibinfo{year}{2019}).
\newblock


\bibitem[\protect\citeauthoryear{Irwin, Sterling, Mysinger, Bolstad, and
  Coleman}{Irwin et~al\mbox{.}}{2012}]%
        {irwin2012zinc}
\bibfield{author}{\bibinfo{person}{John~J Irwin}, \bibinfo{person}{Teague
  Sterling}, \bibinfo{person}{Michael~M Mysinger}, \bibinfo{person}{Erin~S
  Bolstad}, {and} \bibinfo{person}{Ryan~G Coleman}.}
  \bibinfo{year}{2012}\natexlab{}.
\newblock \showarticletitle{ZINC: a free tool to discover chemistry for
  biology}.
\newblock \bibinfo{journal}{\emph{Journal of chemical information and
  modeling}} \bibinfo{volume}{52}, \bibinfo{number}{7} (\bibinfo{year}{2012}),
  \bibinfo{pages}{1757--1768}.
\newblock


\bibitem[\protect\citeauthoryear{Jin, Barzilay, and Jaakkola}{Jin
  et~al\mbox{.}}{2018}]%
        {jin2018junction}
\bibfield{author}{\bibinfo{person}{Wengong Jin}, \bibinfo{person}{Regina
  Barzilay}, {and} \bibinfo{person}{Tommi Jaakkola}.}
  \bibinfo{year}{2018}\natexlab{}.
\newblock \showarticletitle{Junction tree variational autoencoder for molecular
  graph generation}.
\newblock \bibinfo{journal}{\emph{arXiv preprint arXiv:1802.04364}}
  (\bibinfo{year}{2018}).
\newblock


\bibitem[\protect\citeauthoryear{Kingma and Ba}{Kingma and Ba}{2014}]%
        {kingma2014adam}
\bibfield{author}{\bibinfo{person}{Diederik~P Kingma} {and}
  \bibinfo{person}{Jimmy Ba}.} \bibinfo{year}{2014}\natexlab{}.
\newblock \showarticletitle{Adam: A method for stochastic optimization}.
\newblock \bibinfo{journal}{\emph{arXiv preprint arXiv:1412.6980}}
  (\bibinfo{year}{2014}).
\newblock


\bibitem[\protect\citeauthoryear{Kingma and Dhariwal}{Kingma and
  Dhariwal}{2018}]%
        {kingma2018glow}
\bibfield{author}{\bibinfo{person}{Durk~P Kingma} {and}
  \bibinfo{person}{Prafulla Dhariwal}.} \bibinfo{year}{2018}\natexlab{}.
\newblock \showarticletitle{Glow: Generative flow with invertible 1x1
  convolutions}. In \bibinfo{booktitle}{\emph{Advances in Neural Information
  Processing Systems}}. \bibinfo{pages}{10215--10224}.
\newblock


\bibitem[\protect\citeauthoryear{Kobyzev, Prince, and Brubaker}{Kobyzev
  et~al\mbox{.}}{2019}]%
        {kobyzev2019normalizing}
\bibfield{author}{\bibinfo{person}{Ivan Kobyzev}, \bibinfo{person}{Simon
  Prince}, {and} \bibinfo{person}{Marcus~A Brubaker}.}
  \bibinfo{year}{2019}\natexlab{}.
\newblock \showarticletitle{Normalizing flows: Introduction and ideas}.
\newblock \bibinfo{journal}{\emph{arXiv preprint arXiv:1908.09257}}
  (\bibinfo{year}{2019}).
\newblock


\bibitem[\protect\citeauthoryear{Kusner, Paige, and
  Hern{\'a}ndez-Lobato}{Kusner et~al\mbox{.}}{2017}]%
        {kusner2017grammar}
\bibfield{author}{\bibinfo{person}{Matt~J Kusner}, \bibinfo{person}{Brooks
  Paige}, {and} \bibinfo{person}{Jos{\'e}~Miguel Hern{\'a}ndez-Lobato}.}
  \bibinfo{year}{2017}\natexlab{}.
\newblock \showarticletitle{Grammar variational autoencoder}. In
  \bibinfo{booktitle}{\emph{Proceedings of the 34th International Conference on
  Machine Learning-Volume 70}}. JMLR. org, \bibinfo{pages}{1945--1954}.
\newblock


\bibitem[\protect\citeauthoryear{Landrum et~al\mbox{.}}{Landrum
  et~al\mbox{.}}{2006}]%
        {landrum2006rdkit}
\bibfield{author}{\bibinfo{person}{Greg Landrum} {et~al\mbox{.}}}
  \bibinfo{year}{2006}\natexlab{}.
\newblock \bibinfo{title}{RDKit: Open-source cheminformatics}.
\newblock
\newblock


\bibitem[\protect\citeauthoryear{Liu, Kumar, Ba, Kiros, and Swersky}{Liu
  et~al\mbox{.}}{2019}]%
        {liu2019graph}
\bibfield{author}{\bibinfo{person}{Jenny Liu}, \bibinfo{person}{Aviral Kumar},
  \bibinfo{person}{Jimmy Ba}, \bibinfo{person}{Jamie Kiros}, {and}
  \bibinfo{person}{Kevin Swersky}.} \bibinfo{year}{2019}\natexlab{}.
\newblock \showarticletitle{Graph normalizing flows}. In
  \bibinfo{booktitle}{\emph{Advances in Neural Information Processing
  Systems}}. \bibinfo{pages}{13556--13566}.
\newblock


\bibitem[\protect\citeauthoryear{Liu, Allamanis, Brockschmidt, and Gaunt}{Liu
  et~al\mbox{.}}{2018}]%
        {liu2018constrained}
\bibfield{author}{\bibinfo{person}{Qi Liu}, \bibinfo{person}{Miltiadis
  Allamanis}, \bibinfo{person}{Marc Brockschmidt}, {and}
  \bibinfo{person}{Alexander Gaunt}.} \bibinfo{year}{2018}\natexlab{}.
\newblock \showarticletitle{Constrained graph variational autoencoders for
  molecule design}. In \bibinfo{booktitle}{\emph{Advances in Neural Information
  Processing Systems}}. \bibinfo{pages}{7795--7804}.
\newblock


\bibitem[\protect\citeauthoryear{Ma, Chen, and Xiao}{Ma et~al\mbox{.}}{2018}]%
        {ma2018constrained}
\bibfield{author}{\bibinfo{person}{Tengfei Ma}, \bibinfo{person}{Jie Chen},
  {and} \bibinfo{person}{Cao Xiao}.} \bibinfo{year}{2018}\natexlab{}.
\newblock \showarticletitle{Constrained generation of semantically valid graphs
  via regularizing variational autoencoders}. In
  \bibinfo{booktitle}{\emph{Advances in Neural Information Processing
  Systems}}. \bibinfo{pages}{7113--7124}.
\newblock


\bibitem[\protect\citeauthoryear{Madhawa, Ishiguro, Nakago, and Abe}{Madhawa
  et~al\mbox{.}}{2019}]%
        {madhawa2019graphnvp}
\bibfield{author}{\bibinfo{person}{Kaushalya Madhawa},
  \bibinfo{person}{Katushiko Ishiguro}, \bibinfo{person}{Kosuke Nakago}, {and}
  \bibinfo{person}{Motoki Abe}.} \bibinfo{year}{2019}\natexlab{}.
\newblock \showarticletitle{GraphNVP: An Invertible Flow Model for Generating
  Molecular Graphs}.
\newblock \bibinfo{journal}{\emph{arXiv preprint arXiv:1905.11600}}
  (\bibinfo{year}{2019}).
\newblock


\bibitem[\protect\citeauthoryear{Mullard}{Mullard}{2017}]%
        {mullard2017drug}
\bibfield{author}{\bibinfo{person}{Asher Mullard}.}
  \bibinfo{year}{2017}\natexlab{}.
\newblock \showarticletitle{The drug-maker's guide to the galaxy}.
\newblock \bibinfo{journal}{\emph{Nature News}} \bibinfo{volume}{549},
  \bibinfo{number}{7673} (\bibinfo{year}{2017}), \bibinfo{pages}{445}.
\newblock


\bibitem[\protect\citeauthoryear{Papamakarios, Nalisnick, Rezende, Mohamed, and
  Lakshminarayanan}{Papamakarios et~al\mbox{.}}{2019}]%
        {papamakarios2019normalizing}
\bibfield{author}{\bibinfo{person}{George Papamakarios}, \bibinfo{person}{Eric
  Nalisnick}, \bibinfo{person}{Danilo~Jimenez Rezende}, \bibinfo{person}{Shakir
  Mohamed}, {and} \bibinfo{person}{Balaji Lakshminarayanan}.}
  \bibinfo{year}{2019}\natexlab{}.
\newblock \showarticletitle{Normalizing Flows for Probabilistic Modeling and
  Inference}.
\newblock \bibinfo{journal}{\emph{arXiv preprint arXiv:1912.02762}}
  (\bibinfo{year}{2019}).
\newblock


\bibitem[\protect\citeauthoryear{Parmar, Vaswani, Uszkoreit, Kaiser, Shazeer,
  Ku, and Tran}{Parmar et~al\mbox{.}}{2018}]%
        {parmar2018image}
\bibfield{author}{\bibinfo{person}{Niki Parmar}, \bibinfo{person}{Ashish
  Vaswani}, \bibinfo{person}{Jakob Uszkoreit}, \bibinfo{person}{{\L}ukasz
  Kaiser}, \bibinfo{person}{Noam Shazeer}, \bibinfo{person}{Alexander Ku},
  {and} \bibinfo{person}{Dustin Tran}.} \bibinfo{year}{2018}\natexlab{}.
\newblock \showarticletitle{Image transformer}.
\newblock \bibinfo{journal}{\emph{arXiv preprint arXiv:1802.05751}}
  (\bibinfo{year}{2018}).
\newblock


\bibitem[\protect\citeauthoryear{Paul, Mytelka, Dunwiddie, Persinger, Munos,
  Lindborg, and Schacht}{Paul et~al\mbox{.}}{2010}]%
        {paul2010improve}
\bibfield{author}{\bibinfo{person}{Steven~M Paul}, \bibinfo{person}{Daniel~S
  Mytelka}, \bibinfo{person}{Christopher~T Dunwiddie},
  \bibinfo{person}{Charles~C Persinger}, \bibinfo{person}{Bernard~H Munos},
  \bibinfo{person}{Stacy~R Lindborg}, {and} \bibinfo{person}{Aaron~L Schacht}.}
  \bibinfo{year}{2010}\natexlab{}.
\newblock \showarticletitle{How to improve R\&D productivity: the
  pharmaceutical industry's grand challenge}.
\newblock \bibinfo{journal}{\emph{Nature reviews Drug discovery}}
  \bibinfo{volume}{9}, \bibinfo{number}{3} (\bibinfo{year}{2010}),
  \bibinfo{pages}{203}.
\newblock


\bibitem[\protect\citeauthoryear{Popova, Shvets, Oliva, and Isayev}{Popova
  et~al\mbox{.}}{2019}]%
        {popova2019molecularrnn}
\bibfield{author}{\bibinfo{person}{Mariya Popova}, \bibinfo{person}{Mykhailo
  Shvets}, \bibinfo{person}{Junier Oliva}, {and} \bibinfo{person}{Olexandr
  Isayev}.} \bibinfo{year}{2019}\natexlab{}.
\newblock \showarticletitle{MolecularRNN: Generating realistic molecular graphs
  with optimized properties}.
\newblock \bibinfo{journal}{\emph{arXiv preprint arXiv:1905.13372}}
  (\bibinfo{year}{2019}).
\newblock


\bibitem[\protect\citeauthoryear{Ramakrishnan, Dral, Rupp, and
  Von~Lilienfeld}{Ramakrishnan et~al\mbox{.}}{2014}]%
        {ramakrishnan2014quantum}
\bibfield{author}{\bibinfo{person}{Raghunathan Ramakrishnan},
  \bibinfo{person}{Pavlo~O Dral}, \bibinfo{person}{Matthias Rupp}, {and}
  \bibinfo{person}{O~Anatole Von~Lilienfeld}.} \bibinfo{year}{2014}\natexlab{}.
\newblock \showarticletitle{Quantum chemistry structures and properties of 134
  kilo molecules}.
\newblock \bibinfo{journal}{\emph{Scientific data}}  \bibinfo{volume}{1}
  (\bibinfo{year}{2014}), \bibinfo{pages}{140022}.
\newblock


\bibitem[\protect\citeauthoryear{Rogers and Hahn}{Rogers and Hahn}{2010}]%
        {rogers2010extended}
\bibfield{author}{\bibinfo{person}{David Rogers} {and} \bibinfo{person}{Mathew
  Hahn}.} \bibinfo{year}{2010}\natexlab{}.
\newblock \showarticletitle{Extended-connectivity fingerprints}.
\newblock \bibinfo{journal}{\emph{Journal of chemical information and
  modeling}} \bibinfo{volume}{50}, \bibinfo{number}{5} (\bibinfo{year}{2010}),
  \bibinfo{pages}{742--754}.
\newblock


\bibitem[\protect\citeauthoryear{Schlichtkrull, Kipf, Bloem, Van Den~Berg,
  Titov, and Welling}{Schlichtkrull et~al\mbox{.}}{2018}]%
        {schlichtkrull2018modeling}
\bibfield{author}{\bibinfo{person}{Michael Schlichtkrull},
  \bibinfo{person}{Thomas~N Kipf}, \bibinfo{person}{Peter Bloem},
  \bibinfo{person}{Rianne Van Den~Berg}, \bibinfo{person}{Ivan Titov}, {and}
  \bibinfo{person}{Max Welling}.} \bibinfo{year}{2018}\natexlab{}.
\newblock \showarticletitle{Modeling relational data with graph convolutional
  networks}. In \bibinfo{booktitle}{\emph{European Semantic Web Conference}}.
  Springer, \bibinfo{pages}{593--607}.
\newblock


\bibitem[\protect\citeauthoryear{Shi, Xu, Zhu, Zhang, Zhang, and Tang}{Shi
  et~al\mbox{.}}{2020}]%
        {shi2020graphaf}
\bibfield{author}{\bibinfo{person}{Chence Shi}, \bibinfo{person}{Minkai Xu},
  \bibinfo{person}{Zhaocheng Zhu}, \bibinfo{person}{Weinan Zhang},
  \bibinfo{person}{Ming Zhang}, {and} \bibinfo{person}{Jian. Tang}.}
  \bibinfo{year}{2020}\natexlab{}.
\newblock \showarticletitle{GraphAF: a Flow-based Autoregressive Model for
  Molecular Graph Generation}.
\newblock \bibinfo{journal}{\emph{ICLR 2020, Addis Ababa, Ethiopia, Apr.26-Apr.
  30, 2020}} (\bibinfo{year}{2020}).
\newblock


\bibitem[\protect\citeauthoryear{Simonovsky and Komodakis}{Simonovsky and
  Komodakis}{2018}]%
        {simonovsky2018graphvae}
\bibfield{author}{\bibinfo{person}{Martin Simonovsky} {and}
  \bibinfo{person}{Nikos Komodakis}.} \bibinfo{year}{2018}\natexlab{}.
\newblock \showarticletitle{Graphvae: Towards generation of small graphs using
  variational autoencoders}. In \bibinfo{booktitle}{\emph{International
  Conference on Artificial Neural Networks}}. Springer,
  \bibinfo{pages}{412--422}.
\newblock


\bibitem[\protect\citeauthoryear{Sun, Zhao, Gilvary, Elemento, Zhou, and
  Wang}{Sun et~al\mbox{.}}{2019}]%
        {sun2019graph}
\bibfield{author}{\bibinfo{person}{Mengying Sun}, \bibinfo{person}{Sendong
  Zhao}, \bibinfo{person}{Coryandar Gilvary}, \bibinfo{person}{Olivier
  Elemento}, \bibinfo{person}{Jiayu Zhou}, {and} \bibinfo{person}{Fei Wang}.}
  \bibinfo{year}{2019}\natexlab{}.
\newblock \showarticletitle{Graph convolutional networks for computational drug
  development and discovery}.
\newblock \bibinfo{journal}{\emph{Briefings in bioinformatics}}
  (\bibinfo{year}{2019}).
\newblock


\bibitem[\protect\citeauthoryear{Weininger, Weininger, and Weininger}{Weininger
  et~al\mbox{.}}{1989}]%
        {weininger1989smiles}
\bibfield{author}{\bibinfo{person}{David Weininger}, \bibinfo{person}{Arthur
  Weininger}, {and} \bibinfo{person}{Joseph~L Weininger}.}
  \bibinfo{year}{1989}\natexlab{}.
\newblock \showarticletitle{SMILES. 2. Algorithm for generation of unique
  SMILES notation}.
\newblock \bibinfo{journal}{\emph{Journal of chemical information and computer
  sciences}} \bibinfo{volume}{29}, \bibinfo{number}{2} (\bibinfo{year}{1989}),
  \bibinfo{pages}{97--101}.
\newblock


\bibitem[\protect\citeauthoryear{You, Liu, Ying, Pande, and Leskovec}{You
  et~al\mbox{.}}{2018}]%
        {you2018graph}
\bibfield{author}{\bibinfo{person}{Jiaxuan You}, \bibinfo{person}{Bowen Liu},
  \bibinfo{person}{Zhitao Ying}, \bibinfo{person}{Vijay Pande}, {and}
  \bibinfo{person}{Jure Leskovec}.} \bibinfo{year}{2018}\natexlab{}.
\newblock \showarticletitle{Graph convolutional policy network for
  goal-directed molecular graph generation}. In
  \bibinfo{booktitle}{\emph{Advances in Neural Information Processing
  Systems}}. \bibinfo{pages}{6410--6421}.
\newblock


\bibitem[\protect\citeauthoryear{Zhavoronkov, Ivanenkov, Aliper, Veselov,
  Aladinskiy, Aladinskaya, Terentiev, Polykovskiy, Kuznetsov,
  et~al\mbox{.}}{Zhavoronkov et~al\mbox{.}}{2019}]%
        {zhavoronkov2019deep}
\bibfield{author}{\bibinfo{person}{Alex Zhavoronkov}, \bibinfo{person}{Yan~A
  Ivanenkov}, \bibinfo{person}{Alex Aliper}, \bibinfo{person}{Mark~S Veselov},
  \bibinfo{person}{Vladimir~A Aladinskiy}, \bibinfo{person}{Anastasiya~V
  Aladinskaya}, \bibinfo{person}{Victor~A Terentiev}, \bibinfo{person}{Daniil~A
  Polykovskiy}, \bibinfo{person}{Maksim~D Kuznetsov}, {et~al\mbox{.}}}
  \bibinfo{year}{2019}\natexlab{}.
\newblock \showarticletitle{Deep learning enables rapid identification of
  potent DDR1 kinase inhibitors}.
\newblock \bibinfo{journal}{\emph{Nature biotechnology}} \bibinfo{volume}{37},
  \bibinfo{number}{9} (\bibinfo{year}{2019}), \bibinfo{pages}{1038--1040}.
\newblock


\end{thebibliography}
}

\newpage
\appendix

\end{document}